\documentclass{applemlr}

\usepackage{amsmath}
\usepackage{enumerate}
\usepackage{algorithm}
\usepackage{algpseudocode}
\usepackage{amsfonts}
\usepackage{amsthm}
\usepackage{cleveref}
\usepackage{diagbox}
\usepackage{colortbl}
\usepackage{amssymb}
\usepackage{xspace}
\usepackage{wrapfig}
\usepackage{adjustbox}
\usepackage{tabularx}
\usepackage{booktabs}
\usepackage{mathtools}
\usepackage{tikz}
\usepackage{enumitem}
\usepackage{silence}
\usepackage{dsfont}
\usepackage[table]{xcolor}
\usepackage[dvipsnames]{xcolor}
\usepackage{multirow}
\usepackage{makecell}
\usepackage{xfakebold}

\usepackage{amsmath,amsfonts,bm}

\def\eqref#1{equation~\ref{#1}}

\def\1{\bm{1}}

\DeclareMathAlphabet{\mathsfit}{\encodingdefault}{\sfdefault}{m}{sl}
\SetMathAlphabet{\mathsfit}{bold}{\encodingdefault}{\sfdefault}{bx}{n}

\definecolor{textgray}{HTML}{6E6E73}
\usetikzlibrary{positioning, calc}
\usetikzlibrary{decorations.pathmorphing}

\makeatletter
\patchcmd{\wrong@fontshape}{\@gobbletwo}{}{}{}
\makeatother
\numberwithin{equation}{section}
\makeatletter
\AtBeginDocument{
  \urlstyle{sf}
  
}
\makeatother

\usepackage{xcolor}
\definecolor{AppleWhite}{RGB}{255,255,255}
\definecolor{ApplePrimaryCoolGray}{RGB}{116,128,139}
\definecolor{AppleCoolGray1}{RGB}{199,209,214}
\definecolor{AppleCoolGray2}{RGB}{147,174,190}
\definecolor{AppleCoolGray3}{RGB}{124,147,160}
\definecolor{AppleCoolGray4}{RGB}{92,102,109}
\definecolor{AppleCoolGray5}{RGB}{78,93,100}
\definecolor{AppleCoolGray6}{RGB}{53,60,65}
\definecolor{AppleBlack}{RGB}{0,0,0}
\definecolor{AppleSecondaryChartGray}{RGB}{168,168,168}
\definecolor{AppleChartGray2}{RGB}{233,233,233}
\definecolor{AppleChartGray3}{RGB}{211,211,211}
\definecolor{AppleChartGray4}{RGB}{190,190,190}
\definecolor{AppleChartGray5}{RGB}{140,140,140}
\definecolor{AppleChartGray6}{RGB}{102,102,102}
\definecolor{AppleChartGray7}{RGB}{64,64,64}
\definecolor{ApplePrimaryChartBlue}{RGB}{84,151,193}
\definecolor{AppleBlue2}{RGB}{212,229,239}
\definecolor{AppleBlue3}{RGB}{169,202,223}
\definecolor{AppleBlue4}{RGB}{127,177,209}
\definecolor{AppleBlue5}{RGB}{71,130,166}
\definecolor{AppleBlue6}{RGB}{55,99,128}
\definecolor{AppleBlue7}{RGB}{45,72,89}
\definecolor{ApplePrimaryChartGreen}{RGB}{83,172,121}
\definecolor{AppleGreen2}{RGB}{212,234,221}
\definecolor{AppleGreen3}{RGB}{169,213,188}
\definecolor{AppleGreen4}{RGB}{126,193,155}
\definecolor{AppleGreen5}{RGB}{58,140,82}
\definecolor{AppleGreen6}{RGB}{39,102,54}
\definecolor{AppleGreen7}{RGB}{29,58,31}
\definecolor{ApplePrimaryChartYellow}{RGB}{253,195,93}
\definecolor{AppleYellow2}{RGB}{254,240,214}
\definecolor{AppleYellow3}{RGB}{254,224,174}
\definecolor{AppleYellow4}{RGB}{254,210,134}
\definecolor{AppleYellow5}{RGB}{230,168,69}
\definecolor{AppleYellow6}{RGB}{191,131,46}
\definecolor{AppleYellow7}{RGB}{153,107,54}
\definecolor{ApplePrimaryChartOrange}{RGB}{250,151,92}
\definecolor{AppleOrange2}{RGB}{254,229,214}
\definecolor{AppleOrange3}{RGB}{252,203,173}
\definecolor{AppleOrange4}{RGB}{252,178,133}
\definecolor{AppleOrange5}{RGB}{227,121,68}
\definecolor{AppleOrange6}{RGB}{191,87,46}
\definecolor{AppleOrange7}{RGB}{143,59,36}
\definecolor{ApplePrimaryChartRed}{RGB}{227,94,105}
\definecolor{AppleRed2}{RGB}{248,215,217}
\definecolor{AppleRed3}{RGB}{241,174,180}
\definecolor{AppleRed4}{RGB}{234,135,143}
\definecolor{AppleRed5}{RGB}{196,63,77}
\definecolor{AppleRed6}{RGB}{153,35,53}
\definecolor{AppleRed7}{RGB}{102,19,43}
\definecolor{ApplePrimaryChartPurple}{RGB}{161,150,204}
\definecolor{ApplePurple2}{RGB}{231,228,242}
\definecolor{ApplePurple3}{RGB}{208,202,229}
\definecolor{ApplePurple4}{RGB}{185,176,217}
\definecolor{ApplePurple5}{RGB}{128,113,171}
\definecolor{ApplePurple6}{RGB}{89,76,128}
\definecolor{ApplePurple7}{RGB}{62,46,101}
\definecolor{AppleCoolGray}{RGB}{116,128,139}
\definecolor{AppleChartGray}{RGB}{168,168,168}
\definecolor{AppleBlue}{RGB}{84,151,193}
\definecolor{AppleGreen}{RGB}{83,172,121}
\definecolor{AppleYellow}{RGB}{253,195,93}
\definecolor{AppleOrange}{RGB}{250,151,92}
\definecolor{AppleRed}{RGB}{227,94,105}
\definecolor{ApplePurple}{RGB}{161,150,204}

\definecolor{light}{RGB}{125, 125, 125}
\crefname{tcb@cnt@pbox}{code}{code}
\Crefname{tcb@cnt@pbox}{Code}{Code}
\crefname{assumption}{assumption}{assumption}
\Crefname{assumption}{Assumption}{Assumptions}

\usepackage{fvextra}
\newtcolorbox[auto counter]{pbox}[2][]{
  colback=white,
  title=Code~\thetcbcounter: #2,
  #1,fonttitle=\sffamily,
  fontupper=\sffamily,
  arc=2pt,
  colframe=bgcolor,
  coltitle=fgcolor,
  colbacktitle=bgcolor,
  toptitle=0.25cm,
  bottomtitle=0.125cm
}

\tcbuselibrary{skins, breakable}
\newtcolorbox{tldrbox}{
    enhanced,
    breakable,
    colback=AppleOrange2,
    colframe=AppleOrange2,
    coltext=AppleOrange7,
    boxrule=0pt,
    arc=10pt,
    width=\linewidth,
    left=10pt, right=10pt,
    top=8pt, bottom=8pt,
}
\let\cref\Cref
\newcounter{prompt}

\usepackage{titletoc}
\titlecontents{section}
  [2.2em]                                   %
  {\bfseries}                             %
  {\contentslabel{2em}}                   %
  {}                                      %
  {\hfill\contentspage}                   %
  [\vspace{0.75ex}]                       %

\usepackage{wrapfig}

\makeatletter
\newcommand\applefootnote[1]{%
  \begingroup
  \renewcommand\thefootnote{}%
  \renewcommand\@makefntext[1]{\noindent##1}%
  \footnote{#1}%
  \addtocounter{footnote}{-1}%
  \endgroup
}
\makeatother

\definecolor{cverbbg}{gray}{0.90}

\title{RLTLDR: Self-improvement by Internalizing Self-generated Feedback}

\author{Michael Kirchhof}
\author{Eleonora Gualdoni}
\author{Andrew Szot}
\author{Khashayar Gatmiry}
\author{Aryo Lotfi}
\author{Abbas Kazerouni}
\author{Omar Attia}
\author{Sanjoy Chowdhury}
\author{Alexander Toshev}

\affiliation{Apple}

\abstract{The common paradigm of reinforcement learning with verifiable rewards (RLVR) is to let agents make multiple attempts at a task, and optimize towards the successful ones.
This becomes problematic in the realms of self-improvement, where tasks are so difficult that the agent has a low or even no chance of success, and where there are no teacher models or example solutions to distill from.
In this paper, we introduce RLTL;DR. %
After each failed attempt, we show the policy the verifier outputs and let it write its own feedback, in the form of a single TL;DR insight. The next rollout is conditioned on all previous insights, and we sequentially sample rollouts until a solution is found. Moreover, we enable backpropagation on the in-context insights to internalize a direct task $\rightarrow$ insight mapping.
On challenging tool-calling and coding
datasets (filtered to Pass@128=0), standard GRPO training of a Qwen 3.5 9B Thinking policy stays flat at a Pass@1 of 0\% to 1\%.
RLTL;DR breaks through this learning barrier, achieving a Pass@1 of 14--31\% with insights in context during training and, crucially, 12--13\% when no insight is in context at eval time.
We identify that the key is the task $\rightarrow$ insight internalization.
To study this further, we reduce our approach to SFTL;DR, training only on (task, insight) tuples, without showing or backpropagating on any rollouts.
Training on only 4k of these tuples recovers almost the full performance of RLTL;DR and classical SFT on full rollouts. This demonstrates a promising compacted training paradigm of the form "on this sort of task, keep this sort of thing in mind", which we hope to inspire future research on. %
}

\date{\sffamily\today}

\usepackage[textwidth=18mm]{todonotes}
\newcommand{\method}{RLTL;DR\xspace}

\begin{document}

\maketitle

\section{Introduction}

Reinforcement learning with verifiable rewards (RLVR, \citeauthor{lambert2024tulu}, \citeyear{lambert2024tulu}) presents an LLM agent policy with a task and lets the policy attempt to solve it in multiple parallel rollouts. The rewards are then checked for correctness via a verifier (unit tests and final state checks). Successful rollouts receive a positive reward and unsuccessful ones a negative one, for example via GRPO \citep{shao2024deepseekmath,liu2025understanding,yu2026dapo}, improving the agent's performance over time. %

This paradigm fails when tasks are so hard that the policy does not produce any successful rollouts in 128 attempts. Starved of learning signal, the training loop does not take off. Further, in self-improvement scenarios such as agentic coding, we assume that the agent is at the frontier; there is no stronger teacher model or golden example solution to learn from. The policy has to explore the problem step by step and internalize findings from its own attempts.

\begin{figure*}
    \centering
    \includegraphics[width=\linewidth, trim=1cm 8cm 2.7cm 1.3cm, clip]{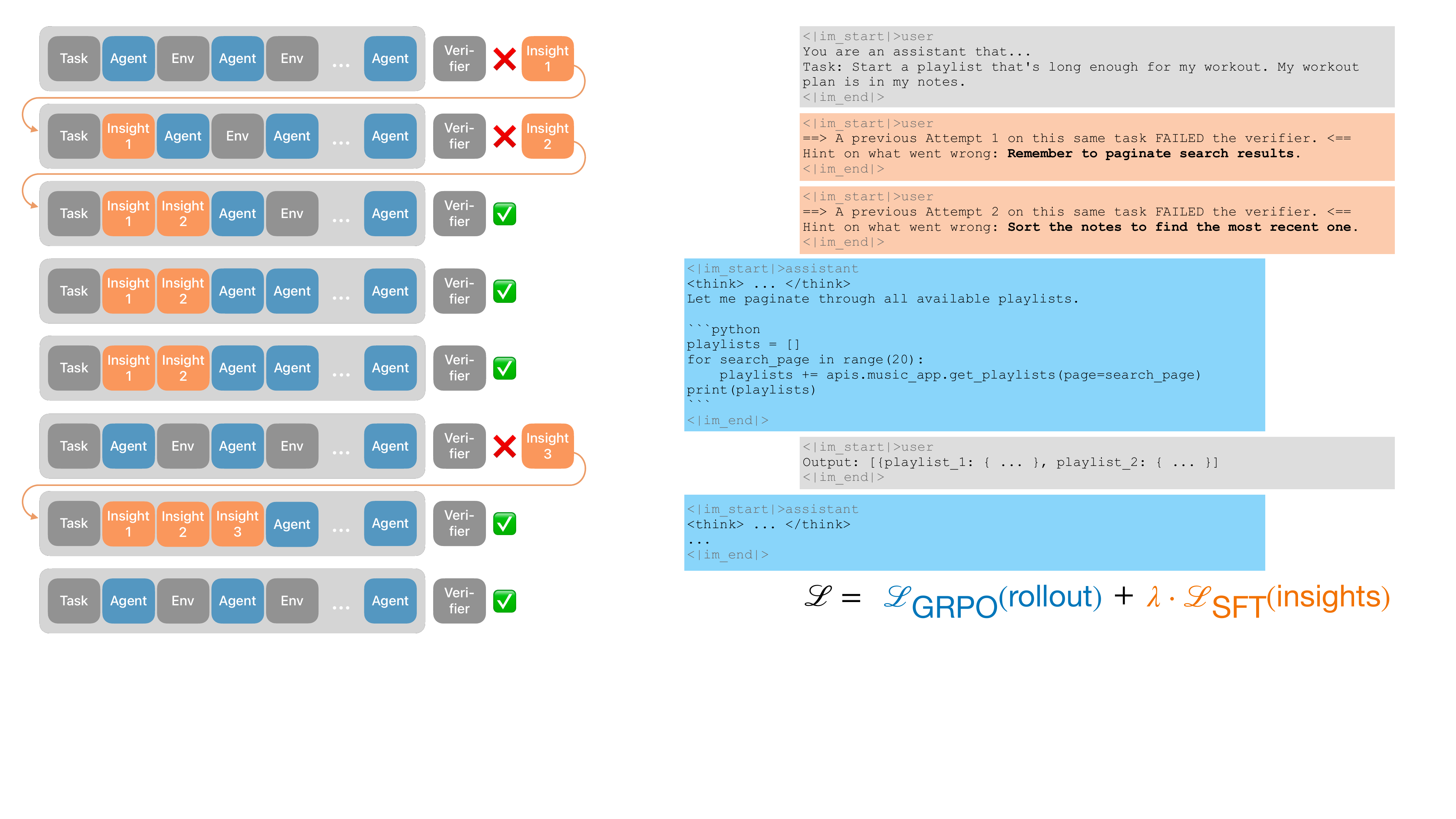}
    \caption{RLTL;DR rolls out attempts sequentially. When an attempt fails, we let the policy generate its own feedback based on the verifier outcomes, in the form of a high-level TL;DR insight. This insight is fed into next attempts to search for solutions more efficiently, especially on very hard tasks. When the running average success rate of the task is high enough, like in the 6th and 8th rollout, we attempt the task without insights. Besides a GRPO loss on the rollouts, we backpropagate the insights (the bold words on the right) into the model via an SFT loss. Though never predicted at test time, this internalizes and generalizes these higher-level learnings.}
    \label{fig:fig1}
\end{figure*}

We propose two techniques that, in combination, allow breaking through the learning barrier: First, we improve the exploration by moving from parallel i.i.d. rollouts to sequential rollouts. After each attempt, the verifier code is run to check for success, and if the attempt failed, the policy itself is handed the previous attempt and the error message to produce feedback, in the form of a higher-level TL;DR insight like \emph{"Remember to paginate search results."}. In the next rollout, the policy is conditioned on the task and the previous TL;DR learnings in context. We find that this sequential sampling lifts the exploration phase of RLVR out of the zone of no successful signals, finding at least one solution for Pass@k$=$14--59\% of the tasks. However, at test time, when the policy has to provide a solution at the first attempt, without any insights in context, it stays close to its original performance. 

We thus make a minimal second change to the update phase of the model: We activate the backpropagation mask on the self-generated insight tokens that are in the context ($\approx$17 tokens per insight). %
This trains a \emph{task} $\rightarrow$ \emph{something to keep in mind} mapping into the policy that, although never generated at test time since insights are inserted in the form of user messages, internalizes the findings and generalizes them to similar problems through sheer smoothness of backpropagation. %

We find that this breaks through the learning barrier in %
agentic and coding benchmarks on tasks where the Qwen 3.5 9B Thinking \citep{qwen35} fails for 128 attempts. While GRPO stays flat at 0\% to 1\%, our RLTL;DR achieves 11--14\% Pass@1 (at eval time; without any insights or sequential sampling). %
We ablate the training and find that the key ingredient is indeed the backpropagation on the \emph{task} $\rightarrow$ \emph{insight} mapping. Even if we turn off the GRPO loss (and thus all backpropagation on tokens from the rollout), the SFT loss on the insight tokens alone achieves almost full performance. %

\section{Related Works}

\paragraph{The learning barrier of RLVR.}
The standard RLVR recipe samples a group of i.i.d.\ rollouts per task, scores them with a verifier, and updates with a group-relative advantage
\citep{lambert2024tulu,shao2024deepseekmath,yu2026dapo,liu2025understanding}.
This has a structural failure mode: when every rollout in a group receives the same reward, either all are correct or all are incorrect, the advantage vanishes and the gradient is zero (a phenomenon variously named advantage collapse, the learning cliff, or exploration inefficiency
\citep{xia2026learninghintreinforcementlearning,agrawal2026ocgrpo,agashe2026cbrl}.
When all rollouts are correct, one can apply different loss functions or mitigations such as entropy control, pass@$k$ objectives, or difficulty-matched curricula \citep{mahrooghi2026goldilocks,chen2025passk}. %
But on the frontier splits we target, success rates are close to zero, deprived of any learnable signal.

\paragraph{Guiding exploration with privileged information.} A growing body of work manufactures at least one success by injecting information the policy will not have at test time. These approaches differ along three axes: the \emph{source} of the guidance, its \emph{explicitness}, %
and the \emph{transfer mechanism} to ensure the policy still performs when no guidance is available at evaluation time. Sources range from gold solutions and stronger teacher models
\citep{zhang2025stephint,zhu2025scafgrpo,agrawal2026ocgrpo} to the policy's own failed attempts \citep{nvidia2026igrpo,song2026rltf,szot2026sge}. Explicitness ranges from a verbatim prefix of the reference solution \citep{agrawal2026ocgrpo,zhang2025stephint} to a single conceptual pointer
\citep{nudging2025}. %
Transfer mechanisms range from none at all \citep{nvidia2026igrpo,zhang2025stephint}, i.e., trusting that improvements under the guided prompt carry over to the bare one, through mixing guided and unguided rollouts within the same group \citep{nudging2025,agashe2026cbrl}, to importance-sampling corrections that make the gradient unbiased for the unguided objective \citep{agrawal2026ocgrpo}. In this paper, we assume there is no external source of guidance except failed unit tests, i.e., the policy needs to find improvements itself. We further aim to simplify both the insight itself and the transfer mechanism as much as possible. RLTL;DR notes down short self-generated insights during training, and simply backpropagates them to internalize the task $\rightarrow$ insight mapping and generalize it to similar tasks.

\paragraph{Internalizing context into weights.}
Training a model to behave as if a context was present when it is not is called \emph{context distillation} \citep{askell2021general,snell2022distillingcontext}, recently revisited on-policy \citep{tsinghua2026contextreturns}. Closest to us are SDPO \citep{hubotter2026sdpo} and RLTF \citep{song2026rltf}, which turn feedback into a self-distillation signal over the rollout. We do not distill the full rollout: we predict the $\approx$17 insight tokens. %
This relies on the recent finding that models can apply knowledge in a different format than trained on \citep{shrivastava2026echo,cook2026programming,lu2026policy,nakkiran2026trained}, going from a short insight to generating a full rollout, enabling to internalize skills and experience \citep{meituan2026skill0,seed2026,ruc2026experienceinternalization}. %

\section{Methods}

\subsection{RL and GRPO preliminaries}

We focus on multi-step agentic tasks where a large language model (LLM) agent interacts with an environment to achieve a goal.
We formalize this as a partially observable Markov decision process (POMDP) with a goal space $\mathcal{G}$, observation space $\mathcal{O}$, and an action space $\mathcal{A}$, all of which are in natural language, and a binary reward function $R$. 
The LLM policy $\pi$ generates an action $a_t \sim \pi(\cdot | h_t)$ including a think trace and a code block, given the chat history $h_t = (g, a_1, o_1, \dots, o_{t-1})$ that starts with the task goal $g \in \mathcal{G}$, followed by agent messages $a \in \mathcal{A}$, and the output that their code produces in $o \in \mathcal{O}$. The episode ends at step $T$ when the agent emits an end-of-task action or reaches the maximum episode horizon of 50 actions. The code verifier adds a final observation $o_T$ and a binary reward $r$. We denote this full trajectory $\tau=(h_T, a_T, o_T, r)$. 
Our objective is to train the agent $\pi_\theta$, parameterized by $\theta$, to optimize the expected outcome reward $r$.
In the environments we consider, the per-step observation $o_t$ offers rich information about the agent's interactions with the environment.
For example, $o_t$ shows search results that the agent printed in its code block $a_t$, or tracebacks if the code execution failed. The final observation $o_T$ includes failed asserts and is only visible to the feedback generator below.

We start from a common reinforcement learning (RL) setup for LLM agents using Group Relative Policy Optimization (GRPO) \citep{shao2024deepseekmath}.
For a task $g \in \mathcal{G}$, GRPO samples $K$ trajectories $\{\tau_k\}_{k=1}^{K}$ in parallel and normalizes the rewards into advantages $\hat{A}_k=r_k-\sum_{i=1}^K r_i$. 
The action likelihoods are off-policy corrected, since $\theta$ has already evolved from its version $\theta_\text{old}$ that collected rollouts, multiplied with the advantages, and clipped with $\epsilon=0.2$ to give the GRPO loss:
\begin{equation}
\mathcal{L}_{\text{GRPO}}(\theta)
=
\mathbb{E}_{\{\tau_k\sim\pi_\text{old}\}_{k=1}^K, t=1, \dotsc, T(k)}
\left[
\min\left(
\frac{\pi_\theta(a_t|h_t)}{\pi_{\theta_\text{old}}(a_t|h_t)}\hat{A}_k,\,
\operatorname{clip}_\epsilon\left(\frac{\pi_\theta(a_t|h_t)}{\pi_{\theta_\text{old}}(a_t|h_t)}\right)\hat{A}_k
\right)
\right]
\label{eq:grpo}
\end{equation}
RL then iterates phases of sampling rollouts given the current policy on several tasks, and updating the policy with the collected rollouts and $\mathcal{L}_{\text{GRPO}}(\theta)$. Since we work mostly on very hard tasks, we stabilize the training with some enhancements from literature, such as dropping division by standard deviation in the above advantages, and our own, which we detail in \Cref{app:loss_details}. 

\subsection{Sequential self-generated insights}
We introduce \method as a new RL training method for solving training tasks that are extremely challenging for the LLM agent, and where neither a stronger teacher agent nor example solutions are available.
On such challenging problems, the LLM agent is unlikely to succeed through random sampling, causing repeated attempts to all receive zero outcome rewards and thus provide no learning signal for the RL training objective in \cref{eq:grpo} \citep{yue2025does,wu2025invisible}.

The first component of \method addresses this problem by modifying the GRPO sampling phase so that the agent attempts the same task multiple times in a row, conditioned on self-generated insights from previous attempts.
Specifically, after attempting the problem, the agent can reflect on its attempt and use information from the environment observations and failed unit tests (in $o_T$) to determine how to improve the subsequent attempt.
We call this natural language assessment of what should be improved in the next attempt ``insight''.

For a given task, the policy generates the first trajectory as usual, with $a_t \sim \pi_\theta(\cdot | h_t)$.
After it finishes the rollout $\tau_1$, if it failed, it self-generates a short insight text $f_1 \sim \pi_\theta(\cdot | \tau_1)$.
The insight generation prompt asks the agent to think, summarize, analyze errors, and finally output the insight $f_i$ as a single-sentence summary of what to improve on the next attempt (\Cref{app:feedback_generation}).
We proceed with generating the next attempt on the task. Whenever at the $k$-th attempt $\leq50\%$ of the attempts $1, \dotsc, k-1$ are successful, we insert all insights collected so far. We add the insights $\{f_i\}_{i=1}^I$ from the previous $I \leq k-1$ failed attempts as additional chat messages $\tilde{h}_t=(g, f_1,  \dotsc, f_I, \dotsc)$ after the goal, generating the next attempt with them in context via $a_t \sim \pi_\theta(\cdot | \tilde{h}_t)$. This sequential generation continues for $K$ attempts. We use the $50\%$ boundary to insert insights only on tasks where the agent is struggling, to maintain a goldilocks zone of success rates \citep{mahrooghi2026goldilocks}. \Cref{sec:abl_losses} shows that \method is robust to the choice of the heuristic. 

\method generates insight using the policy itself, so with the same (evolving) weights $\theta$.
As we later demonstrate, this successive insight and retry mechanism enables the model to solve more challenging tasks than the base model alone, other prompting approaches, or exploration approaches. To compare fairly, we match GRPO's and our number of attempts per task.
We treat the $K$ successive attempts as a single GRPO group.
While the rollouts are conditioned on different (or no) insights, we find no performance differences when splitting advantage groups (\Cref{sec:abl_losses}) and prefer simplicity.

In synchronous rollout collection, one could expect sequential sampling to take $K\times$ longer than parallel GRPO sampling, plus the cost of generating the insight. But with asynchronous rollout collection with continuous batching and caching, at $K=8$ we observe the sampling phase to be $4.5\times$ slower. Since update phases and other fixed costs stay equal, the overall walltime increases by $1.5\times$. This is of course not important to begin with in very difficult settings where GRPO simply fails to learn. %
One could also increase the number of tasks that are rolled out in parallel during rollout collection by $K\times$ to alleviate any throughput differences and ensure maximum GPU utilization. We do not do this in this paper in order to give GRPO and RLTL;DR the same amount of data per update phase, for benchmarking fairness.

\subsection{Insight internalization} \label{sec:feedback internalization}
While training with sequential insights improves the policy's ability to explore solutions during training, we find that learning to solve tasks with insights in the context does not directly transfer to solving tasks without insights in the context.
This is significant because, at test time, the agent must succeed in a single attempt. The second component of \method internalizes the self-generated insights so that the performance gains from sequential sampling with insights (during training) transfer to operation without insight (during evaluation).

\method overcomes this issue by introducing a new self-distillation objective that trains the model to connect insights directly to the task.
We train the LLM to predict the insights $\{f_i\}_{i=1}^I$ it generated in previous rollouts (and has in context later attempts) using the task description $g$ alone as input $\pi_\theta(f_1, \dotsc, f_I | g)$. This internalizes the knowledge "on this sort of task, keep these sort of things in mind", with generalization to similar tasks happening thanks to semantic smoothness \citep{nakkiran2026trained}. %
We implement this objective as a standard supervised fine-tuning (SFT) loss for next-token prediction.
We denote this SFT loss by $\mathcal{L}_{\text{SFT}}(\theta)$ and add it to the GRPO loss to obtain the final \method training objective $\mathcal{L} = \mathcal{L}_\text{GRPO} + \lambda \mathcal{L}_\text{SFT}$, where $\lambda$ is the insight internalization strength.
We show in \Cref{sec:grpo_epoch_benefit} that \method is robust to the choice of $\lambda$ and that $\lambda=0.5$ is a good default.

For example, in \cref{fig:fig1} the agent has gathered two insights from previous failed attempts. In the third attempt, it has them in context as two additional user chat messages.  $\mathcal{L}_\text{SFT}$ backpropagates to increase the log likelihoods of the tokens 
\texttt{"Remember to paginate search results."} 
given the context \texttt{"<|im\_start|>\\user\textbackslash{}nYou are an assistant that... Task: Start a playlist that's long enough for
my \\
workout. My workout plan is in my notes.\textbackslash{}n<|im\_end|>\textbackslash{}n<|im\_start|>user\textbackslash{}n==> A previous \\
Attempt 1 on this same task FAILED the verifier. <==\textbackslash{}nHint on what went wrong: "}. It backpropagates \texttt{"Sort the notes to find the most recent one."} the same way, conditional on the task, first insight, and start of the second insight message. Note that $\pi_\theta(f_1, f_2|g)$ is part of the chat history $\tilde{h}_3$ anyways, hence the log likelihoods are already computed in the RL update phase. So, practically, $\mathcal{L}_\text{SFT}$ is simply implemented by editing the backpropagation mask of the context of the third attempt, without increasing runtime.

Some important distinctions between \method and prior work are that $\mathcal{L}_\text{SFT}$ in \method is used solely to internalize the insights about this (and similar) tasks.
Other work \citep{song2026rltf} trains on $\pi_\theta(f|\tau)$, i.e., improving the policy's capability to self-critique given an attempt. We find this to underperform compared to \method direct prediction of insights from the instruction alone (\Cref{sec:abl_feedback_gen}).
Further, in \method, $\mathcal{L}_\text{SFT}$ is used solely as a means to internalize the insights and change behavior on the task.
\method never \emph{generates} insight via the $\pi_\theta(f|g)$ we backpropagate on, neither during training, where the inserted insight comes from analyzing previous failed attempts, nor at test time, where the agents needs to one-shot solutions without any insight or sequential attempts.
Last, \method does not run out of context budget because each insight inserted into the context averages $17$ tokens.
We provide full implementation details in \Cref{app:reproduction}.

\section{RLTL;DR breaks through the learning barrier}

\subsection{Experiment setup}
\textbf{Datasets. } In Appworld \citep{trivedi-etal-2024-appworld}, the policy has to retrieve information and conduct state-changing actions on a simulated device via multi-step tool-calling. It has 90 train, 57 dev, 168 test-normal, and 417 test-challenge tasks. 
Since this dataset is relatively small (especially after filtering them to very hard splits below), we also use a proprietary dataset similar to Appworld, but with 16k train tasks, that we call Synthetic-API (SAPI). %
For more general coding, we use 2641 Leetcode problems \citep{xia2025leetcodedataset}.  

The self-improvement scenarios that we aim to study are characterized by tasks that are so hard that the models are struggling to find solutions even with high budgets. To emulate this difficulty, we subsample the above datasets: We use tasks where our policy, Qwen 3.5 9B (with thinking), has no successes in 128 attempts, i.e., Pass@128=0. This filters down SAPI to 458 tasks. For the smaller Appworld dataset, we combine the train, dev, and test-normal splits (and keep test-challenge unseen), leaving 34 tasks after filtering to the Pass@128=0 set. Leetcode %
has 123 remaining tasks. %

\textbf{Baselines. } We compare RLTL;DR to three baselines. RLTF-SD \citep{song2026rltf} is a recent method, similar in kind. It uses a Self Distillation loss to train a rollout generated with insight into the policy without insight. For fairness, we provide it with the same sequential rollouts and insight generation strategies as for our method. Second, we compare against Strategy-guided Exploration (SGE, \citeauthor{szot2026sge}, \citeyear{szot2026sge}). This aims to explore more solutions by prepending summaries of previous failed or successful attempts on a task (though without insights on what went wrong), and prompting the model to try something else. %
Finally, we compare against a standard GRPO baseline. %
We tune our GRPO baseline extensively. Our Qwen 3.5 9B GRPO baseline trained only on Appworld-train achieves 72.2\% Pass@1 on test-challenge. As of the release of this paper, the best agent on the official Appworld leaderboard is a frontier model in a custom harness, at 73.4 Pass@1.

\textbf{Deconfounded evaluation. } Since some of the rollouts are conditioned on insights in their context during training, we deconfound our metrics. Our curves and metrics, both during train and eval, \emph{always show the performance on rollouts without insight in context} (and are macro-averaged across all tasks, see \cref{app:deconfounded_eval}).  %
This allows to compare fairly, treating insight only as a train aid. %

We use the official heldout sets for evaluation, without filtering for hard tasks, to show ensure the policies do not degradate outside hard tasks. Appworld test-challenge has 417 new tasks on both the seven seen and two new apps. SAPI has 1624 tasks on 4 unseen apps. Leetcode has 228 unseen problems. We take multiple attempts to achieve 2k rollouts for each dataset.

\subsection{Results} \label{sec:exp_frontier}

\begin{figure*}[t]
  \centering
  \begin{subfigure}[b]{0.32\textwidth}
    \includegraphics[width=\textwidth]{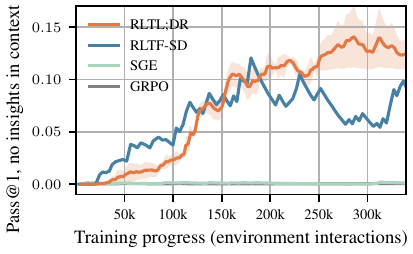}
    \caption{Synthetic API}
  \end{subfigure}
  \hfill
  \begin{subfigure}[b]{0.32\textwidth}
    \includegraphics[width=\textwidth]{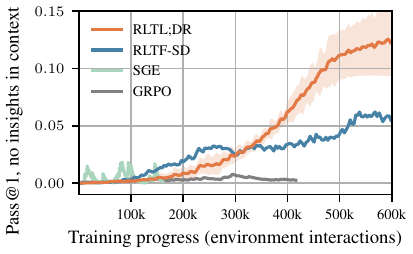}
    \caption{Appworld}
  \end{subfigure}
  \hfill
  \begin{subfigure}[b]{0.32\textwidth}
    \includegraphics[width=\textwidth]{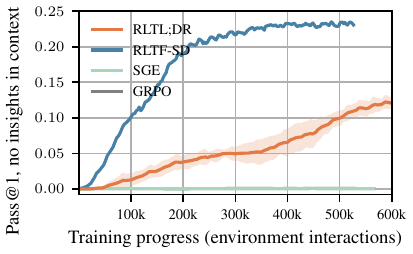}
    \caption{Leetcode}
  \end{subfigure}
  \caption{Pass@1 through training, measured only on rollouts without insights (comparable between approaches). Average and std across 3 seeds. GRPO fails to learn since it is starved of learning signal. RLTL;DR and RLTF-SD break through this learning barrier. Leetcode probably overfit.}
  \label{fig:frontier_difficulty}
\end{figure*}

\cref{fig:frontier_difficulty} shows the policy's Pass@1 throughout training, on rollouts without insights in context. Insights thus only acted indirectly to improve learning signal in previous batches, and performance can be directly compared. On SAPI, every task has been seen once after $\approx$170k environment interactions, and reported performances are before backpropagating on any given task, so that on SAPI, the train curve until $\approx$170k can be seen as eval curves on a rolling basis. Appworld and Leetcode loop every 13k and 9k steps, so we defer to the heldout splits below for judging generalization.

GRPO fails to learn on these very hard tasks, staying flat at 0\% to 1\%. This is because GRPO is starved of successful rollouts and thus learning signal. SGE behaves similarly. Although it conditions on previous attempts and it is highlighted that they failed, it does not reflect on failed unit tests. We observe that this misleads next rollouts (reproducing \citet{cheng2026contextualdrag}, see also \Cref{sec:abl_feedback}). 

\begin{wraptable}{r}{0.45\textwidth}
\small
\vspace{-4mm}
    \centering
    \begin{tabular}{lrrr}
    \toprule
         & SAPI & Appworld & Leetcode \\
    \midrule
       GRPO  & 57.8\% & 33.7\% & 55.1\% \\
       SGE & 57.7\% & 33.5\% & 56.4\% \\
       RLTF-SD & 53.0\%& 38.9\% & 42.2\% \\
       RLTL;DR & 91.1\% & 61.7\% & 49.1\% \\
    \bottomrule
    \end{tabular}
    \caption{Pass@1 on heldout eval sets after training on the Pass@128=0 splits. RLTL;DR's internalized insights generalize, to unseen and also easier tasks. These numbers should not be cited as benchmark scores as their train data is only a subset of tasks and includes Appworld test-normal.}
    \label{tab:heldout}
    \vspace{-6mm}
\end{wraptable}

RLTL;DR, on the other hand, breaks through the learning barrier and reaches a Pass@1 of 12--13\%. As can be seen from the SAPI curve before 170k steps, and the heldout splits in \Cref{tab:heldout}, the internalization generalizes insights to new tasks. We do not see this on Leetcode. Both RLTL;DR and RLTF-SD find solutions during training (GRPO does not) but seem to overfit in the process. We discuss this in \Cref{sec:dis}. When including rollouts with insights in context, train-time Pass@1 is 14--31\%, and train-time Pass@k is 14--59\%, demonstrating how RLTL;DR finds learning signals on previously impossible tasks.

RLTF-SD also learns. We refrain from claims on either approach outperforming. Instead, we see RLTL;DR and RLTF-SD as two promising approaches of acquiring off-policy knowledge (the same knowledge, since in our experiments they both use our sequential sampling and insight generation pipeline). But while RLTF-SD learns by backpropagating examples, RLTL;DR learns from the high-level insight. These two complementary backpropagation signals can be combined by simply changing the gradient mask in RLTF-SD. We observe performance gains with this in \Cref{app:rltfsd_lsft}.

We also train on a slightly easier split of Synthetic API with a baseline Pass@1$=$4\% in \Cref{app:sapi_4pct}, with equivalent observations. On the unfiltered datasets in \Cref{sec:exp_fulldata}, where only $\leq$3--5\% of tasks are frontier-difficult, GRPO is able to learn, and RLTL;DR neither helps nor hurts performance (except Leetcode). We thus see RLTL;DR as a method for training on challenging tasks.

\begin{tldrbox}
\textbf{TL;DR: } RLTL;DR breaks through the learning barrier that GRPO faces on  Pass@128=0 tasks, reaching 12-13\% Pass@1 (evaluated without insights in context or sequential attempts). %
\end{tldrbox}

\section{Reducing to the secret sauce: From RLTL;DR to SFTL;DR} \label{sec:sftldr}

In the development of RLTL;DR, internalizing the insight via $\mathcal{L}_\text{SFT}$ was the switch that enabled self-improvement on very hard tasks. In this section, we reduce to only $\mathcal{L}_\text{SFT}$, and make the perhaps surprising finding that we can learn only from insights, without full rollouts.

\subsection{Experiment setup}
\paragraph{Dataset. } We dedicate the remainder of this paper to SAPI, due to its sheer size. We use a split containing the 458 frontier-difficult tasks, plus 184 very difficult tasks as explained in \Cref{app:sapi_4pct}. Qwen 3.5 9B achieves 4\% Pass@1 on this split. The 642 tasks are split into 256 eval and 386 train tasks. GRPO training still fails, while RLTL;DR achieves 21.5\% Pass@1.

\paragraph{RLTL;DR ablations. } Starting with the standard RLTL;DR run (GRPO loss and SFT loss on insight with strength $\lambda=0.5$), we first reduce and then fully deactivate $\lambda$. Then, on the contrary, we deactivate the GRPO loss and train only with the SFT loss. These trainings are all online, so rollouts are collected as the policy improves.

\paragraph{Standard SFT. } We also train on all successful rollouts collected throughout the standard RLTL;DR run via offline SFT. These SFT runs use a log likelihood loss on the agent actions in the rollouts, with insights in the contexts but without SFT loss on the insights, over 10 epochs. %
Besides Pass@1 on train and heldout tasks, we track the number of tokens we backpropagate on, as well as the number of tokens we need to forward calculate to generate the context KV caches. We train on different amounts of SFT data to be able to compute-match the SFTL;DR results.

\paragraph{SFTL;DR. } In the runs named SFTL;DR, we use the same rollouts but only apply the SFT loss on the insight tokens $\pi_\theta(f_1, \dots, f_I|g)$, without backpropagating (or even forward calculating) the actual rollouts, which would come autoregressively after the insight. The standard SFTL;DR setup backpropagates on all insights in each rollout, possibly multiple times (insights appear in multiple rollouts per task, and in multiple combinations). In SFTL;DR-deduplicated, we further simplify this training. We create unique tuples $(g, f)$ of the insights per task, 4k in total, and then train $\pi_\theta(f|g)$ one-by-one. This resembles a training where we only train "on this task, remember this insight".

\subsection{Reducing RLTL;DR to just SFTL;DR} 

\begin{table}[t]
    \caption{We find that $\mathcal{L}_\text{SFT}$ drives most performance, on the SAPI frontier difficulty split separated into train and heldout eval tasks. The first four runs are RLTL;DR ablations, showing that the SFT loss on insights is the driving factor. The next are normal SFT train runs on rollouts collected in the first RLTL;DR run. Most rollouts have insight in context, but it is not backpropagated on. The last group of experiments is the simplified SFTL;DR training that only backpropagates on the insights, not the rollouts. Forward tokens and backward tokens concern the logits that need to be computed in the update phase. Rollout collection is another 1B tokens. All results are $\pm0.6\%$ standard deviation.}
    \label{tab:sftldr}
    \footnotesize
    \centering
    \resizebox{1.0\textwidth}{!}{
    \begin{tabular}{lrrrr}
    \toprule
  & Forward Tokens & Backward Tokens & Train Pass@1 & Eval Pass@1 \\
 \midrule
  RLTL;DR & $~720$M & $12$M & 21.5\% & 18.9\% \\
  RLTL;DR, $\lambda=0.01$ & $~720$M & $12$M & 9.3\% & 14.0\%  \\
  RLTL;DR, $\lambda=0$ & $~720$M & $11$M & 6.1\% & 13.2\% \\ 
  RLTL;DR, no GRPO loss, only $\mathcal{L}_\text{SFT}$ & $9.2$M & $839$k & 20.0\% & 17.0\% \\
\midrule
  SFT on all 3.5k full rollouts & $400$M & $6.6$M & 28.0\% & 20.9\%\\
  SFT on 1k rollouts & $107$M & $1.9$M & 27.1\% & 21.0\%\\
  SFT on 100 rollouts & $9.5$M & $217$k & 25.9\% & 20.8\%\\
\midrule
  SFTL;DR on insights & $2.9$M & $292$k & 17.8\% & 16.8\% \\
  SFTL;DR on insights, deduplicated & $4.6$M & $68$k & 19.1\% & 16.9\% \\
\bottomrule
    \end{tabular}
    }
\end{table}

\Cref{tab:sftldr} shows the Pass@1 on the train and unseen eval tasks, both evaluated without insight in context or sequential sampling. The first four runs show that $\mathcal{L}_\text{SFT}$ is the driving factor in RLTL;DR. Reducing its mixture weight from $\lambda=0.5$ to $\lambda=0.01$ or $0$ severly impacts both train and eval performance (increasing it beyond $\lambda > 0.5$ did not further improve it). While the model might be learning how to solve tasks given insight, it does not internalize the insight itself to one-shot solve tasks once insight is not available in context. In fact, entirely removing the GRPO loss and only utilizing the SFT loss on insight tokens (but still in the RL setup of interleaved rollout and update phases) recovers almost the full performance of RLTL;DR. Nevertheless, the additional learning signal from the full rollouts lets runs with non-zero GRPO loss converge faster (\Cref{sec:grpo_epoch_benefit}), so we recommend leaving it activated. %

Classical SFT on the rollouts collected during the RLTL;DR is slightly above RLTL;DR's performance, though still close to standard deviation. Indeed, we can reduce the number of train rollouts from 3.5k to 100 without losing much performance, as performance scales sub-linearly with the amount of input. This gives a compute-matched baseline to compare to SFTL;DR. 

Interestingly, simply training on (task, insight) tuples in SFTL;DR, without ever seeing the insight "in action" in a rollout, like above almost reaches SFT and RLTL;DR performance on full rollouts. This might be the most striking result: The model is able to internalize and generalize the knowledge given in an entirely different format from how it will have to put it into code at evaluation time. 
We discuss this finding in the light of recent findings on the surprising smoothness of training of large language models in \cref{sec:dis_sftldr}.

A final remark is that training only on (task, one-sentence insight) tuples also reduces the train compute. Its 4.6M forward (context) and 68k backwards (insight) tokens approach the performance of SFT on 100 rollouts with 9.5M forward and 217k backward tokens, and SFT on 3.5k rollouts with 400M forward and 6.6M backward tokens. We underline that this is purely a reduction in policy update compute. Full rollouts still need to be collected in order to generate the insights, which is the largest block of about $\approx$1B tokens in all approaches.

\begin{tldrbox}
\textbf{TL;DR: } LLMs can be trained simply via a $\mathcal{L}_\text{SFT}$ loss on (task, one-sentence insight) tuples, without backpropagating on any rollouts, to internalize and generalize high-level knowledge.
\end{tldrbox}

\section{What makes for good insights?} \label{sec:abl_feedback}

The insight we provide the model is short and procedural: typically, a single sentence (17 words on average), %
like \emph{``You need to mark the article as read instead of just viewing it''}, see \Cref{app:examples}. %
Crucially, the insight need not be too specific: as we show below, this level of abstraction is key for reaching good learning signal. We ablate multiple insight design choices on the SAPI-frontier split: how detailed it is, how it is generated, and how many iterations of insight are sequentially attached.

\textbf{Insight content.} We vary how detailed the insights are that are placed into the agent's context. Instead of the single-sentence TL;DR, we provide a full diagnostic paragraph on why the attempt failed, optionally a summary of the attempt preceding the diagnostic paragraph, or both plus a proposed code correction. These artifacts are already generated in the main method as a byproduct when giving a TL;DR insight (we just see them as autoregressive generation aids and drop everything except the TL;DR insight), so they give the same hint, just in different level of detail.

\textbf{Insight generator.} First, we degrade the insights by turning off thinking during insight generation, or by not showing failed unit tests. Next, we improve insights by using a separate, more capable teacher model, GLM 5.2 \citep{glm5team2026glm5vibecodingagentic}, in both thinking and non-thinking modes. 

\textbf{Amount of insight.} We set the maximum number of insights in the context to 1, 2, 4, and 8, instead of the default 16. Note that we always use (and backpropagate) the most recent insight.

We train and evaluate like in \Cref{sec:sftldr}. Table~\ref{tab:feedback-ablation} shows that replacing the TL;DR format with more detailed insights hurts performance: train Pass@1 reduces from $21.4$ to $20.3$ with the diagnostic paragraph, and more sharply to $13.8$ when the summary is added, and to $15.6$ when the corrected code is added as well.
This drop is not because detailed insights are less useful in context: As everywhere in this paper, the numbers above measure performance \textit{without} insights in context. %
When we instead measure Pass@1 \textit{with} insights in context, summary + diagnostic paragraph yields the largest benefit of any configuration tested, adding $+41.1\%$ over the Pass@1 of unaided rollouts, against $+36.0\%$ for TL;DR. %
This suggests that detailed hints help the agent solve the task at hand but do not provide a learning signal that can be internalized for the unaided setup, or generalized to other tasks, while TL;DR insights state reusable rules. %
This is in line with expectations in literature, for two reasons: First, detailed insights might include session-specific details that are hard to predict from the goal alone, like IDs (see \citet{lu2026policy} and \Cref{app:example_formats}), preventing internalization due to label noise. Second, even when a detailed insight can be internalized, it may be too specific to transfer to other tasks, as discussed by \citet{xia2026learninghintreinforcementlearning}.

Changing who generates the TL;DR insight matters less than what information the generator has. A stronger teacher gives a modest gain. With 16 insights, GLM~5.2 with thinking reaches $24.5$ vs.\ $21.4$ for the student. With a single insight, the gap grows to about $5$ points ($26.2$ vs.\ $21.4$). Thinking makes little difference for either generator, it slightly increases the train Pass@1 and yet slightly decreases the evaluation Pass@1. Some of these differences are within run-to-run noise: we read them as trends rather than a clear ranking. The same conclusion applies to our analysis of the number of insights (see~Appendix \ref{app:inference_curves} for a complementary study of this design choice at inference time). In contrast, removing access to the failed unit tests has a large effect: the student fails to extract useful insights, and Pass@1 drops to $3.4$. What matters most is thus whether the insight correctly identifies the failure. RLTL;DR seems to handle insight well whether it is generated on- or off-policy (or potentially by humans). In practice, any reasonably capable insight generator works, which can be achieved even with a small model if given access to privileged information, or by stronger generators when available.

\begin{table}
\centering
\caption{Pass@1 (without insight in context) after training with different forms of insight.
$n_{\text{fb}}$ is the maximum number of insights placed in the context.}
\label{tab:feedback-ablation}
\resizebox{1.0\textwidth}{!}{
\begin{tabular}{llrrr}
\toprule
 & Setting & $n_{\text{fb}}$ & Train Pass@1 & Eval Pass@1 \\
\midrule
 & Default (TL;DR format, self-generated with thinking) & 16 & 21.4\% & 19.2\% \\
\midrule
\multirow{3}{*}{Insight content}
 & Diagnostic paragraph                              & 16 & 20.3\% & 18.6\%\\
 & Summary + Diagnostic paragraph                    & 16 & 13.8\% & 16.0\% \\
 & Summary + Diagnostic paragraph + Corrected code   & 16 & 15.7\% & 17.3\% \\
\midrule
\multirow{6}{*}{Insight generator}
 & Student, non-thinking                                           & 16 & 20.1\% & 19.8\% \\
 & Student, no information about failed unit tests (only success)  & 16 & \phantom{0}3.4\% & 10.0\%\\
 & GLM 5.2, non-thinking & 16 & 21.0\% & 20.5\% \\
 & GLM 5.2, thinking     & 16 & 24.5\% & 20.2\% \\
 & GLM 5.2, non-thinking & 1  & 25.5\% & 22.7\% \\
 & GLM 5.2, thinking     & 1  & 26.2\% & 20.3\% \\
\midrule
\multirow{4}{*}{Amount of insight}
 & 1 insight  & 1 & 20.6\% & 18.9\% \\
 & 2 insights & 2 & 21.9\% & 19.3\% \\
 & 4 insights & 4 & 19.1\% & 17.8\% \\
 & 8 insights & 8 & 23.3\% & 20.3\% \\
\bottomrule
\end{tabular}

}
\end{table}

\begin{tldrbox}
\textbf{TL;DR: } Insights can be generated by any model and should use any privileged information that makes them accurate. But they should remain broad enough to generalize to other tasks.
\end{tldrbox}

\section{Discussion} \label{sec:dis}

\subsection{The surprising learning only from (task, insight) tuples} \label{sec:dis_sftldr}

The "secret sauce" of our approach seems to be training on the insight tokens given only the task description. Although these tokens are never produced at eval time, this training seems to be sufficient to backpropagate the higher-level findings of the insights into the model parameters and implicitly apply them during rollout generation. 

There are three recent works that observe similar phenomena. ECHO \citep{shrivastava2026echo} train an agentic policy that interacts with a terminal, and during the update phase also backpropagate on the terminal outputs. They find that this improves the general knowledge of the policy about the terminal. Like in our setup, these tokens are inside user messages, not in agent messages, and hence the knowledge is just internalized but never explicitly generated. \citet{lu2026policy} make a similar finding in (text-based) embodied agent and search tasks. \citet{cook2026programming} first train the model on code documentations and then test its ability to write code for new tasks. They observe that this knowledge transfers between the two formats, like in our jump from \emph{task} $\rightarrow$ \emph{insight} backpropagation to \emph{task} $\rightarrow$ \emph{rollout} generation. Earlier, \citet{hsieh2023distilling} have noted that backpropagating think traces into LLMs, even if the LLM is used without think traces at test time, improves performance.

The generalization dynamics that drive this transfer are currently unknown. We attribute the effectiveness of (task, insight) training to smoothness during the backpropagation. Our best understanding is that during backpropagation, LLMs act as semantic similarity machines, so that parameters for not just the literal task but semantically similar tasks and different output formats are updated, thanks to having trained on vast amounts of similar tasks and smoothing out their semantic similarities \citep{nakkiran2026trained}.

\subsection{Limits of (task, insight) learning}

Based on this understanding, we expect that compacted (task, insight) training will not work in tasks that are overly specific and share little common rules. For example, if a mathematical proof requires finding a very specific trick, the sequential sampling might help explore this more quickly, but the (task, insight) training will not help generalize to other tasks. We suspect the similar effect to hold for our Leetcode results. We also hypothesize that (task, insight) training is an emergent capability that works only if the model has already been pretrained on enough full rollouts, in order to have a sufficiently smooth network. Last, we believe that there might be some domains in which finding a good insight requires the same capability level as generating a valid solution in the first place. While we did not observe this in the tool calling and coding benchmarks in this paper (potentially because we have the privileged information of the verifier), it might become problematic in domains like automated scientific research. There, internalizing insight might still work as good as learning from full rollouts, but coming up with the insight might be too hard for our sequential rollout strategy.

We do not expect, however, that scaling to larger models would make insight-based learning less effective. On the contrary, we believe that the smoothness that enables learning and generalization from insight is likely to be ever larger in larger models. We also believe that larger models are better insight generators, and potentially able to find errors in a previous attempt even when the verifier outcomes are not revealed to them. We encourage to test this hypothesis in future works.

\subsection{Source and quality of insight}

In \cref{sec:abl_feedback}, we find that performance depends on the insight. Our setup uses the small student model as its own insight generator, but gives access to the verifier code and outcomes as priviledged information. If this is revoked, the insight becomes too low-quality to learn from. Equivalently, larger teacher models generate insight that increases performance further. In preliminary work, we also experimented with not having any LLM generated insight, but just unit tests that output instructive strings, which also gives some performance. 

Our best understanding is that the source of the insight does not matter and can be left as a pragmatic choice. It only matters whether an insight is helpful enough to increase the Pass@k in the next rollout, while being generic enough to transfer to other tasks. We discuss in \cref{app:hint_strength_metrics} that if one does not train insight into the model via an SFT loss but, e.g., a GRPO loss, then other metrics about the insight can become important (such as being not too revealing to get mixed groups of rollouts). 

\subsection{Beyond insight databases}

There are multiple recent works that build databases of insights and use a search system to insert them in the prompt when a similar task comes up \citep{zhang2026memrl,tang2026wikiskill,nasvytis2026core}. We see this as the most promising path when using untrainable (frontier) models. However, if there is the possibility to train the model (or even just an adapter), we believe training insights into the model via $\mathcal{L}_\text{SFT}$ might be the simpler approach. This is because the backpropagation of the insight automatically generalizes it to similar tasks, and the model applies the insights when needed automatically, to the extent that is necessary. This removes the need for a dedicated (and often complex) retrieval system. Of course, performance of this needs to be benchmarked, which we leave as future work.

\section{Conclusion and outlook}

This paper is a first demonstration that directly backpropagating high-level, compacted insights into a policy, conditioned only on the task and not on the rollout, enables the policy to internalize, generalize, and apply these insights. We focus on using this as an auxiliary objective during a self-improvement RLVR loop, which enables breaking through the learning barrier in otherwise learning-signal-starved Pass@128=0 tasks. But we also find that it can be used as a training signal on its own, without requiring full rollouts to backpropagate on. 

This gives rise to multiple next questions: First, how exactly is the knowledge generalized through the backpropagation? We hypothesize this has to do with the smoothness of the parameters of a (sufficiently pretrained) model, implicitly routing the knowledge not just naively to the literal task and literal format of task $\rightarrow$ insight, but to any semantically similar task and output format, including generating a full rollout. Second, where are the limits of this paradigm? Tool-calling might be special in its hard to find but easy to apply insights. We expect that training only on compressed insights is not feasible in all domains, especially in domains where the policy possesses too little pretraining capabilities for the smoothness to emerge, or domains where tasks are so specific that strong enough insights are not applicable to similar problems. Third, which other forms of training become possible if we remove the need for full rollouts? We expect that training only on the insights, without the concrete example, can enable federated learning at scale, learning from feedback or skills written by humans, and learning from summaries of very long rollouts that possibly contain erroneous detours, as is common in self-improvement scenarios.

All tool-use experiments in this work are conducted in research-only simulated environments. The APIs, tasks, datasets, verifier infrastructure, and training procedures described here do not represent or imply any deployed Apple product, production system, or product roadmap.

\applefootnote{ \textcolor{textgray}{\sffamily Apple and the Apple logo are trademarks of Apple Inc., registered in the U.S. and other countries and regions.}}

\bibliographystyle{abbrvnat}
\bibliography{main}

\newpage
\appendix
\crefalias{section}{appendix}
\crefalias{Section}{Appendix}

\startcontents[appendix]
\printcontents[appendix]{}{1}{\section*{Appendix Contents}\vspace{5mm}}
\newpage

\section{Prompts and details of RLTL;DR}

\subsection{Insight generation} \label{app:feedback_generation}

We let the policy generate its own insight after each rollout. After each rollout that the verifier code flags as failed, we take the full rollout chat, and insert it inside a judge prompt in a new context. We use a new conversation rather than continuing the previous chat to reduce contextual drag \citep{cheng2026contextualdrag}. As shown in the prompt below, we insert task, agent actions, observations (shortened if exceedingly long), and unit test outputs by the verifier code. 

We let the agent first think (with a budget of 4096 tokens), then produce a summary, (longer) feedback, a step in which the agent went wrong, a correction, and finally a one-sentence insight. We extract this via json parsing and only use the final one-sentence insight in the paper, the remainder is mostly an autoregressive crux to increase test-time compute before providing the insight. We find that providing the long summary and feedback as hint for the next rollout does not outperform in \Cref{sec:abl_feedback}. In fact, it reduces performance, likely because one-sentence insights are more general and internalizing them via backpropagation can transfer the knowledge to other similar tasks. 

\begin{tcolorbox}[colback=gray!5, colframe=black, fontupper=\small\ttfamily, title={Prompt~\refstepcounter{prompt}\theprompt: Prompt for the insight generation. Note that we let the agent first think, and then generate summary, feedback, wrong step ID, and corrected step, before giving the actual insight. Only the actual insight (in the form of a TL;DR sentence) is used in the majority of the paper, the remainder is generated autoregressively in order to increase accuracy.\label{prompt:feedback}}, breakable,enhanced,before upper={%
        \setlength{\parskip}{0.75\baselineskip}%
        \setlength{\parindent}{0pt}%
        \par
    }]
\begin{Verbatim}[fontsize=\small, breaklines=true, breaksymbolleft={}, breaksymbolright={}]
You are a verifier that is given a step-by-step solution of an agent that has to perform some on-device task for the following task: ...

Here is the step-by-step solution that the agent proposed. At each step, it proposed some code to retrieve information or execute actions, then gets an observation from the environment that executed that code on the simulated device.

============================== AGENT ROLLOUT ==============================
Step 1: 
...
Observation 1:
...
Step 2: 
...
============================== END OF AGENT ROLLOUT ==============================

The verifier code ran and returned this output:
Rollout FAILED the following verifier tests:
Playlist has length 2712 seconds but must be >= 3215 seconds.

As a verifier, you now have to produce a json with FIVE parts: 1) Summarize the agent attempt, 2) Give feedback, 3) Give the step at which it went wrong, 4) Provide the corrected code for that step, 5) Give a one-sentence TL;DR hint.

1) Summarize the attempt
- Give a summary of roughly one paragraph that describes what the agent did.
- You can skip very generic code like API lookups or logins.
- Focus on what the agent searched for and what it edited.
- Also mention what the final answer of the agent was, or which items it changed on device exactly.
- Do NOT consider the failed tests yet, if this was a failed rollout.
- Do NOT critique yet, just summarize.

2) Give feedback
- For successful rollouts, just return an empty string.
- For failed rollouts, look at all failed tests.
- Write a paragraph on what went wrong in the rollout.
- Do NOT reveal what the ground-truth values are (private_data.* and all numbers that are on the right in the unit test asserts). You can reveal what the agent values were (the left values) if it helps explain the error.
- Try to point out where exactly in the code the error is. Stop at the FIRST thing that went wrong.
- Make it independent of the agent rollout, do NOT assume that the rollout will be shown along with your feedback.

3) Give the step at which it went wrong
- Return the ID of the step where the agent produced wrong code.
- If you cannot identify a specific step, output -1.

4) Provide the corrected code for that step
- Return ONLY the python code that the agent should have written for the wrong step in place of what it actually wrote. Not the prior steps, not the following steps.
- The code should be runnable as a drop-in replacement for the buggy step's code.
- Do NOT include explanatory prose inside the code; use code-comments only if essential.
- Do NOT reveal ground-truth values (private_data.* / unit-test RHS). If a value is unknown, query it via the API instead of hard-coding.
- If the rollout was correct or you cannot identify a specific wrong step, return an empty string.

5) Give a one-sentence TL;DR hint
- After creating all the previous four parts, output a single-sentence hint on what went wrong.
- Keep it to ONE sentence, plain language, no code.
- This should be a concise pointer the agent can act on (e.g. "You created the list but gave it the wrong name.").
- For successful rollouts, just return an empty string.
- Do NOT reveal ground-truth values (private_data.* / unit-test RHS).

Output your answer in a JSON. Do NOT output any text except the JSON. The format should be:
{
    "summary": "...",
    "feedback": "...",
    "wrong_step_id": "integer",
    "corrected_step": "...",
    "hint": "..."
}
\end{Verbatim}
\end{tcolorbox}

\subsection{Insight conditioning} \label{app:feedback_conditioning}

Insight conditioning is triggered if there is at least one insights from a previous failure in the current batch (and thus task) and the running average success rate of the batch is $\leq50\%$. We collect all one-sentence insights generated so far in this batch. If the task has already been attempted in an earlier RL update phase, we do not include those insights. As shown in \Cref{fig:fig1}, each insight is inserted as a single user message after the task and before the start of the next rollout. We do not insert the entire previous attempt / chat history, because we found this to introduce contextual drag. 

We track which rollouts are conditioned on insight and which are not, in order to compute the deconfounded metrics in \Cref{app:deconfounded_eval}.

\subsection{Loss details} \label{app:loss_details}

In this section, we describe some changes in the loss function and architecture compared to standard GRPO. We accumulated these changes to improve the performance of the GRPO baseline on Appworld-train. We then use them for all methods for fairness (while possibly giving the baseline a slight advantage due to having tuned the changes towards it, not towards our own method).

\subsubsection{Changes taken over from DAPO}

We utilize changes from literature to improve stability and performance. 

\textbf{KL Divergence. } Just like DAPO \citep{yu2026dapo}, we do not regularize the policy to stay close to the original policy via a KL divergence. We observe that even without KL divergence, the model does not degenerate. 

\textbf{Token-level policy gradients. } DAPO averages the losses on the outside of the sums, rather than within each rollout, so that tokens in shorter rollouts to not obtain a higher implicit weight than tokens in longer rollouts. We do the same.

\textbf{Removing the standard deviation. } Dr. GRPO \citep{liu2025understanding} removes the division by the standard deviation in the standard GRPO advantage function. We also observe improvements when removing it. On very hard tasks, where most samples have a negative gradient, the low standard deviation would otherwise increase the gradients, which increases entropy and destabilizes the model. 

\subsubsection{Positive-ratio filtering} \label{app:positive_ratio_details}

We also add a new trick we call positive-ratio filtering.
In early experiments with the GRPO baseline, we often observed instability due to entropy explosion, followed by policy collapse. We find that on the very hard tasks we train on in our self-improvement setting, the majority of GRPO rewards are negative, thus driving the model to reduce the likelihoods of its known modes and increase its entropy. This entropy increase appears to be rather blind; it pushes the policy away from its current knowledge, but towards no new modes, thus destabilizing training.

Thus, after calculating the advantages of each rollout (and filtering out zero-advantage batches), we filter rollouts to ensure that $75\%$ have a positive reward. 
This trick is primarily intended to stabilize the GRPO baseline. Since it is part of our $\mathcal{L}_\text{GRPO}$ loss, it is also active for RLTL;DR (and the SGE baseline). However, we find that there it is not so critical, since the insight conditioning and SFT loss already lead to many positive-advantage rollouts.

\subsection{Train hyperparameters} \label{app:hyperparams}

Hyperparameters for our train runs are given in \cref{tab:hyperparameters}. We use different hyperparameters for Leetcode and Appworld because Leetcode does a single, long message per task. We deactivate think mode on Leetcode, because on the very hard problems, think mode ran out of token budget. The average number of attempts per task is higher than the minimum number because we use an aynchronous rollout collection, where some workers may take longer to fill their minimum number of attempts (due to working on harder tasks), while other workers continue collecting rollouts. Note also that we follow Qwen's train format, i.e., always only the current action has think tokens in context, previous steps have no think traces. This means that at backprop time, by "minibatch size" we mean a single step (think trace + action), not a full rollout.

All experiments were run on nodes of 8xB200 GPUs, and took 1-4 days of compute each.

\begin{table}   
\caption{Hyperparameters used for train runs.}
    \label{tab:hyperparameters}
    \centering
    \begin{tabular}{lcc}
    \toprule
     & Appworld/SAPI & Leetcode \\
    \midrule
        \multicolumn{3}{c}{Rollout collection phase} \\
    \midrule
        Unique tasks per rollout collection phase & 8 & 128 \\
        Min attempts per task & 8 & 8 \\
        Avg attempts per task (due to async collection) & 21 & 26 \\
        Max steps per attempt & 50 & 1 \\
        Max think tokens per step & 768 & 0 \\
        Max action tokens per step & 512 & 16384 \\
        Temperature & 1 & 1\\
    \midrule
        \multicolumn{3}{c}{Update phase} \\
    \midrule
        Learning rate (const, no warmup) & $3\cdot 10^{-6}$  & $6\cdot 10^{-6}$ \\
        Minibatch size & 4 & 1 \\
        Gradient accumulation steps & 8 & 4\\
        GPUs & 8 & 8\\
        Global batchsize & 256 & 32\\
        Global batchsize (avg tokens) & 102k & 49k \\
        (Global) batches per PPO epoch & 4 & 13 \\
        PPO epochs per update phase & 2 & 2\\
        Backward tokens per update phase & 0.8M & 1.3M \\
    \bottomrule
    \end{tabular}
\end{table}

\subsection{Fixing Qwen's tendency to overthink on hard Leetcode problems} \label{app:qwen_leetcode_fix}

On Leetcode, we notice that Qwen 3.5 9B tends to produce overly long solutions and run out of even very high token budgets of 16k without producing a codeblock. This behavior remains regardless of whether we deactivate thinking (in which case it would reason in the text block), or deactivate thinking and directly open a code block (in which case it would reason in code comments). This only happens on very hard (Pass@128=0) Leetcode tasks, on easier problems it produces code blocks as expected. 

Training RLTL;DR on this split was somewhat trivial; the feedback generator simply always responded to shorten the answer, until (with enough insights of this in the context) the model would comply. We thus seeked out to fix this problem first, so that the insight generation would be more challenging. 

We thus first sample a Pass@128=0 split and GRPO train on it for 100k steps. GRPO quickly picks up the simple first-order statistic that long rollouts have low reward. The final checkpoint almost never runs out of the 16k token budget anymore. We use this checkpoint to start training from on Leetcode experiments in this paper, and the 123 frontier-difficult Leetcode tasks are tasks where this checkpoint still has Pass@128=0. 

\section{Reproducing the RLTL;DR Implementation} \label{app:reproduction}

Reproducing our approach takes three steps. We provide literal-format examples in this paper to verify the implementation.

First, after each rollout, the agent needs to be called again in a new context/chat with the judge prompt (this is important to prevent context drag / bias). This prompt should look as in \Cref{app:feedback_generation}, and we use up to 4096 think and 4096 action tokens for this generation. The output should be a json dict, and the TL;DR insight can be extracted from it programmatically. 

Second, the insight needs to be inserted into context the next time we make a rollout, if the current batch's avg success rate is $\leq50\%$. The insights should be inserted as user messages after the task, \Cref{fig:fig1} gives the exact format including special tokens (only omitting newlines). We found the construction of this context to be a frequent error source and encourage to output the literal context sent to the model during debugging or even as an assert statement during inference. 

Third, in the backpropagation phase of the RL updater, the masks on the insight tokens need to be flipped to backpropagate on the insight tokens. This should look exactly like the example in \Cref{sec:feedback internalization}. Note that when we have multiple insights, we flip the masks of all of them simultaneously, so later insights are learned conditionally on previous ones. This is largely for simplicity (in order to not require constructing new contexts and thus forward computations).

With these three implementations checked, we encourage to first verify that insights work in the given domain by producing a plot like \cref{fig:passatk128}. Then, during training, we encourage to check that the insight advantage described in \Cref{app:hint_strength_metrics} is $>0$. Last, after training, we encourage to check the deconfounded success rate without insights in context as reported throughout this paper and as described in \Cref{app:performance_metrics}.

\section{Examples of generated insights} \label{app:examples}

\subsection{Example TL;DR insights}
\label{app:example_hints}

\newcolumntype{Y}{>{\raggedright\arraybackslash}X}

Figure~\ref{fig:repeated-hints} lists the insights that recur most often across our main run. Each insight names a single corrective action in the vocabulary of the task itself. 

\begin{table*}[h]
    \begin{tabularx}{\linewidth}{
        @{}
        r@{\quad}
        >{\raggedright\arraybackslash}p{0.24\linewidth}
        X
        @{}
    }
    \toprule
    Share & Correction & Examples of insight \\
    \midrule

    15.3\% & Wrong capitalization &
    \begin{itemize}[leftmargin=1.2em, nosep, topsep=0pt]
        \item \emph{Use the uppercase Orange color enum instead of lowercase orange when updating the list color.}
        \item \emph{Use uppercase PM instead of lowercase pm when setting the alarm's AM/PM marker.}
        \item \emph{Use the Priority enum value instead of a string when setting the reminder priority.}
    \end{itemize}
    \\[4pt]

    \midrule

    11.8\% & Unrecorded action &
    \begin{itemize}[leftmargin=1.2em, nosep, topsep=0pt]
        \item \emph{You need to mark the article as read rather than just viewing it.}
        \item \emph{Use the unified search API instead of the ticker search to create a history record that contains the query text.}
        \item \emph{You need to create a download event for the mobile listing instead of using the generic task completion function.}
    \end{itemize}
    \\[4pt]

    \midrule

    11.2\% & Extra parameter &
    \begin{itemize}[leftmargin=1.2em, nosep, topsep=0pt]
        \item \emph{Call \texttt{complete\_task} without passing an answer parameter for non-QA tasks.}
    \end{itemize}
    \\[4pt]

    \midrule

    5.5\% & Unsaved setting &
    \begin{itemize}[leftmargin=1.2em, nosep, topsep=0pt]
        \item \emph{You need to set the default chart horizon preference instead of just viewing a chart.}
        \item \emph{The list sorting preference needs to be saved to the database, not just the sort configuration updated.}
    \end{itemize}
    \\[4pt]

    \midrule

    2.1\% & Over-filtering &
    \begin{itemize}[leftmargin=1.2em, nosep, topsep=0pt]
        \item \emph{Remove the \texttt{route\_type} filter to see all hiking trails instead of getting no results.}
    \end{itemize}
    \\[4pt]

    \midrule

    1.2\% & Argument position &
    \begin{itemize}[leftmargin=1.2em, nosep, topsep=0pt]
        \item \emph{Pass the access token as the first positional argument, not as a keyword argument.}
    \end{itemize}
    \\[4pt]

    \midrule

    52.3\% & Other &
    \begin{itemize}[leftmargin=1.2em, nosep, topsep=0pt]
        \item \emph{You need to confirm the brand ownership is properly recognized before adding items to cart.}
        \item \emph{Use cash instead of checking as the account type.}
        \item \emph{You need to actually read or scroll through the article content before completing the task to create a read history entry.}
        \item \emph{Use the proper alphabetical sort field instead of title when updating the list sort configuration.}
    \end{itemize}
    \\
    \bottomrule

    \end{tabularx}

    \caption{
        Example insight provided by the verifier. Groups of insights pointing to the same correction are identified by lexical overlap.
    }
    \label{fig:repeated-hints}
\end{table*}

\subsection{Examples including previous autoregressive fragments}
\label{app:example_formats}

Table~\ref{tab:feedback_examples} places the three feedback components we ablate side by side on the same failure types. The ablation settings of Section~\ref{sec:abl_feedback} either give the TL;DR insight, or instead the diagnostic paragraph, or instead the diagnostic paragraph and the corrected code. The diagnostic paragraph is accompanied with a summary of the previous attempt, not shown here.

The TL;DR insight names one action to take, whereas the diagnostic paragraph adds the verifier's own logic (which field of the database it reads, which assertion it evaluates, etc.). The corrected code goes further and embeds episode-bound literals, like object identifiers and, in some cases, session credentials lifted verbatim from the rollout, such as access tokens and passwords.  The next episode has a different database, different identifiers and a different token, so none of that detail has an analogue the agent could reuse.

\begin{table}[h]
\caption{The three feedback components on three failure types. Rows are building blocks, not separate conditions: the \emph{Summary + diagnostic paragraph} setting replaces the insight with the paragraph, and \emph{Summary + diagnostic paragraph + corrected code} appends the code to it (summaries omitted for space). Specificity escalates from a transferable rule to episode-bound detail: the hint names an action, the paragraph names the verifier's internal check, and the code carries concrete identifiers such as \texttt{product\_id=1} and, in the last row, an access token valid only for that episode.}
\label{tab:feedback_examples}
\centering\small
    \begin{tabular}{@{}p{0.15\linewidth}p{0.78\linewidth}@{}}
    \toprule
    \multicolumn{2}{@{}l@{}}{\textbf{Failure type: answer argument passed to a non-QA task}}\\
    \midrule
    Diagnostic paragraph & ``The task failed because the agent called \texttt{complete\_task()} with an answer
    parameter, but non-QA tasks should not include an answer. The verifier only checks that the BPM
    filter was set to 125--125, the playlist search results were recorded, and the embed code exists
    in the database --- not the submission method.''\\[3pt]
    Corrected code & \texttt{apis.supervisor.complete\_task()}\\[3pt]
    TL;DR insight & ``Remove the answer parameter from the complete task call.'' \\
    \midrule
    \multicolumn{2}{@{}l@{}}{\textbf{Failure type: item not persisted to \emph{saved for later}}}\\
    \midrule
    Diagnostic paragraph & ``The rollout failed because the product was not properly persisted to the \texttt{saved\_for\_later} collection as verified by the database. The agent called \texttt{save\_for\_later} with only \texttt{product\_id} and \texttt{note}, but did not pass the  \texttt{access\_token} parameter which is required for authentication. The verifier checks the database directly via \texttt{show\_saved\_for\_later(db)}, not API responses.''\\[3pt]
    Corrected code & \texttt{apis.shop.save\_for\_later(product\_id=1,}\\
    & \hspace{1em}\texttt{access\_token=access\_token,}\\
    & \hspace{1em}\texttt{note=`Reviewing for backyard purchase')}\\
     TL;DR insight & ``Ensure the product move to saved for later properly persists the item in the saved for later list for the verifier to detect.''\\
    \midrule
    \multicolumn{2}{@{}l@{}}{\textbf{Failure type: search called without the required access token}}\\[3pt]
    \midrule
    Diagnostic paragraph & ``In Step 6, the agent attempts to search for the `Client Sync' task but fails to include the required \texttt{access\_token} parameter in the \texttt{search\_tasks} API call. The search fails because the access token is not provided to authenticate the request. This prevents the task from being retrieved properly for subsequent operations.''\\[3pt]
    Corrected code & \texttt{apis.to\_do\_list.search\_tasks(}\\
    & \hspace{1em}\texttt{access\_token=`b334de24-3a03-4792-a77b-3a313cb3eeb4',}\\
    & \hspace{1em}\texttt{query=`Client Sync')}\\[3pt]
    TL;DR insight & ``Include the access token when making the search request so it can access your authenticated notes.''\\
    \bottomrule
  \end{tabular}
\end{table}

\section{How to evaluate without confounders} \label{app:deconfounded_eval}

RL training with insight in context, and also generally RL training, has many subtle confounders that can make some approaches look better or worse than others. In this section, we break down how we run evaluate to eliminate those confounders, both for standard RL metrics and for metrics that target the quality/strength of insights.

\subsection{Deconfounded RL train and eval metrics} \label{app:performance_metrics}

\textbf{Pass@1. } The perhaps most important metric in RL train curves and evaluation is the Pass@1, the average success rate across all tasks. There are two confounders here, worker throughput and insight conditioning. 

Worker throughput is a problem that mostly arises in asynchronous RL training. In evaluation, we limit each task to be evaluated an exact amount of times (usually 8). But during training, we let workers collect rollouts for their assigned task until the slowest worker has collected its required minimum GRPO groupsize of 8 rollouts. This means that tasks have different amounts of rollouts, and usually easier tasks (that are solved faster) have more rollouts. Looking at the global average success rate during training is thus biased. We always report macro-averaged Pass@1 (and Pass@k), by first averaging success rates per task, and then across tasks. 

The second confounder comes from insight conditioning: Rollouts with insights in context are naturally higher-performing (otherwise the insights are broken). But insights are not available at test time, so the actual metric we care about is performance without insights in context. Thus, during training we calculate the above macro-average both on all rollouts and also only on rollouts that do not have insights in context. One complication to keep in mind here is that the macro average could become biased again if some tasks don't have any rollouts without insights (usually, those are very hard tasks) or any rollouts with insights (very easy tasks). In our setup, this is not a problem, since the very first rollout per task is always without an insight, and so the macro-average is always over the same tasks. Lastly, we fix the ordering of tasks, allowing to compare across time and across approaches.

\textbf{Pass@k. } Pass@k is an important metric to judge whether a task is within the capabilities of a model, at least with some probability. However, $k$ is not constant, neither across tasks nor across different train runs, due to the varying worker throughput in async rollout collection we described above. So some tasks might have $k=8$ and others $k=23$, and Pass@23 will naturally be higher than Pass@8. 

We thus track two metrics: Pass@8 (constant) and Pass@$k$ (all rollouts per task). This is because both serve different purposes. Pass@8 allows to compare the capabilities of models across different RL runs. Pass@k allows to judge the learning dynamics of a single RL run, to understand if the model still has some learning (or more precisely, exploitation) potential, because there is a reasonable gap between Pass@1 and Pass@k, or whether one needs to lean more into exploration, because Pass@k is too low to begin with. In the main paper, we report whichever metric makes more sense in the given context and comment explicitly on the choice.

\textbf{Choice of the $x$-axis. } Especially in train curves, one needs to decide what to plot metrics against on the x-axis. The same holds for after how much train budget one should end training. Recently, many papers have been using the number of updates (as in the back-and-forth between rollout collection and agent update phases). However, we argue this is largely confounded: Consider one run that just samples more rollouts per task, be it by construction by directly or indirectly increasing the GRPO groupsize, or again due to the async worker throughputs. It will have more rollouts to backpropagate on, and higher chances of finding solutions due to the higher Pass@k. This would go completely undetected if only looking at the number of update phases on the $x$-axis. Another issue is that some policies might sample longer rollouts with more environment steps, again providing more learning signal.

We argue that better candidates for the $x$-axis are the number of steps taken in the environment (in chat setups, that is the number of messages sent), the number of tokens generated, or the overall walltime. In our setup, we choose the number of steps taken in the environment, since walltime differs slightly by hardware and occasional vllm crashes. We track all of these metrics though as secondary metrics and plot them against one another. This makes it easy to detect if one RL run differs considerably from another RL run, so that one can investigate whether its advantage comes purely from this, and how to normalize or restrict it back to allow unconfounded comparisons across runs. For example, we make sure to keep group sizes (rollouts per task) comparably distributed across runs.

\textbf{Other secondary metrics. } Besides these main metrics, we also track the policy's entropy (as a first warning sign of instability), its PPO clip rate when online logits start differing strongly from the offline policies logits (indicating too high effective learning rate), and its PPO clip rate on the very first batch in update phases, before any updates (indicating off-policy drift if beyond pure bfloat16 noise, for example when the vllm collection sampler produces vastly different logits than the logits that are recalculated during the update phase). We do not report these metrics in this paper, but track them to verify the stability of training and best possible performance of all approaches and baselines.

\subsection{Gauging insight strength} \label{app:hint_strength_metrics}

A good insight should help find a solution on the next try, but it should also not be so strong that it makes the task trivial and collapses learning signal. We thus track multiple metrics to judge the quality of insights.

\textbf{Insight advantage. } The most straight-forward metric is to compare the average success rate on rollouts with and without insights. This poses the same caveats as described above for Pass@1: These metrics should be macro-averaged first within and then across tasks (because different tasks might have different amounts of insights), and one needs to make sure that this is done for the same set of tasks (easy tasks might not have any rollouts with insights, depending on how one decides to give insights). After controlling for both of these confounders, the Pass@1 with insights minus the Pass@1 without insights, on all tasks that have both rollouts with and without insights, gives the average insight advantage. This estimate of how much an insight increases the probability of finding a solution to the problem should always stay $>0$, and ideally with quite some margin, but may reduce to 0 as training progresses.

\textbf{Insight Pass@k. } In addition to the average insight advantage, we also report the Pass@2$, \dotsc,$ Pass@16 when sampling with insights in context (the Pass@1 being without any insights because none has been generated yet). We find that tracking this even on the off-the-shelf policy before any training often predicts how well the training will respond to insights.

\textbf{Insight too-easy ratio. } It can, however, be that an insight makes a task too easy to solve. The extreme case here would be that an insight just gives the full solution. We thus also track in how many tasks that have rollouts conditioned on insights \emph{all} rollouts conditioned on insights are correct. This collapses the learning signal of GRPO training, and indicates that insights should be less revealing. It cannot be prevented that this occasionally happens, but we aim to keep this ratio below 20\%. 

However, for the SFT loss on insights it does not matter whether insights make tasks too easy, so this insight too-easy ratio is more important in literature that uses only a GRPO loss, and less important for us than the insight advantage metric.

\textbf{Insight too-hard ratio. } On the flipside, insights can be not helpful (or even misleading). To capture this, we track on how many tasks, that have rollouts conditioned on insights, \emph{none} of the insight-conditioned rollouts are correct. This indicates that insights should be made stronger, for example by giving the insight-generator access to the verifier code to make the insight more precise. It is hard to give a strict target value here, since especially on challenging tasks we expect that a big share of tasks is unsolvable even with the occasional insight, but we aim to use insights that keep this ratio below 50\%. 

\textbf{Insight generalization. } The four previous metrics only track how much an insight helps on the task that the insight was generated for. However, the ideal metric would be to track "how much does this insight help teach the model". Unfortunately, this metric is close to intractable, unless one can afford to run an evaluation on heldout data after every train step. Instead, train curves (at least on SAPI) serve as a natural heldout evaluation on a rolling base: At iteration $i$, we collect rollouts on tasks that have never been seen on iteration $1, \dots, i-1$, based on the policy that has been trained with rollouts and insights from iterations $i, \dotsc, i-1$. Then we update on these tasks and move to the next tasks. Naturally, all eval metrics (like Pass@1) are calculated \emph{before} the update. The steepness of the curve indirectly reveals how much the training with insights generalizes to general knowledge about how to solve tasks. We aim to adhere to this principle by training on a large enough dataset, where we see each task only once before we update on it, like in SAPI. In smaller datasets like Appworld, we explicitly note in the main text at which point the training cycles through an epoch boundary, and report heldout performance at the end of training. 

\section{Additional analyses of the main run}
\label{app:additional_analyses}

\subsection{Distance to original policy}
\label{app:distance}

\begin{table}[htbp]
\caption{Difference of trained checkpoints to original Qwen 3.5 9B checkpoint in parameter space. The first three metrics capture the magnitude of the update, the latter two the concentration (higher = more concentrated on a small set of parameters). Comparisons should only be made within RL runs and within SFT runs, since RL at its bfloat16 precision drops many small parameter updates. }
\label{tab:checkpoint_stats}
\resizebox{\textwidth}{!}{%
\begin{tabular}{llrrrrrr}
\toprule
Method & Train loop & $\|\Delta_\theta\|_1$ & $\|\Delta_\theta\|_2$ &  $\|\Delta_\theta\|_1 >$ bf16 precision & Gini($|\Delta_\theta|$) & top-1\% share of $\Delta_\theta^2$ \\
\midrule
RLTL;DR & RL & 7.4$\cdot10^3$ & 0.79 & 2.11\% & 0.99 & 96.88\% \\
RLTL;DR, $\lambda=0.01$ & RL & 3.2$\cdot10^3$ & 0.44 & 1.58\% & 0.99 & 99.66\% \\
RLTL;DR, $\lambda=0$ & RL & 2.4$\cdot10^3$ & 0.34 & 1.58\% & 0.99 & 99.67\% \\
RLTL;DR, no GRPO loss, only $\mathcal{L}_\text{SFT}$ & RL & 9.3$\cdot10^3$ & 0.89 & 2.51\% & 0.99 & 94.07\% \\
\midrule
SFT on all 3.5k full rollouts & SFT & 1.3$\cdot10^6$ & 36.12 & 30.42\% & 0.87 & 35.47\% \\
SFT on 1k rollouts & SFT & 0.8$\cdot10^6$ & 22.72 & 24.86\% & 0.88 & 36.33\% \\
SFT on 100 rollouts & SFT & 0.2$\cdot10^6$ & 7.47 & 22.10\% & 0.88 & 34.85\% \\
\midrule
SFTL;DR on insight & SFT & 0.038$\cdot10^6$ & 1.83 & 8.88\% & 0.95 & 58.25\% \\
SFTL;DR on insight, deduplicated & SFT & 0.045$\cdot10^6$ & 2.05 & 9.56\% & 0.95 & 54.78\% \\
\bottomrule
\end{tabular}%
}
\end{table}

\Cref{tab:checkpoint_stats} reports how much the final trained checkpoint parameters from \Cref{sec:sftldr} differ from the original Qwen 3.5 9B policy. We report the L1 and L2 norm of $\Delta_\theta = \theta_\text{trained} - \theta_\text{original}$ to gauge the general magnitude of the update, as well as how many parameters have been updated beyond bfloat16 precision. To measure the spread of the the update, we report the Gini index (how non-uniformly the L1 norm of updates is distributed across parameters) and how much of the update energy is concentrated in the 1\% of parameters with the most update energy. In both of these metrics, higher means more concentrated. 

First, there is a large difference in general between RL and SFT train runs. The updates in SFT are two orders of magnitude bigger and much more spread out throughout the network parameters. This is not due to RLTL;DR or to SFTL;DR. It reproduces a finding on training sparsity. Indeed, upon deeper analysis, we confirm \cite{shenfeld2026rl}'s effect that this is most likely due to precision during training. While the SFT pipeline acts in fp32 in many parts, the RL pipeline in many parts uses bf16, and so many very small (possibly noisy) updates fall below the bf16 noise threshold. We encourage further exploration of this effect which appears across multiple papers in future works, but for this paper, just note to only compare within RL or within SFT/SFTL;DR runs. 

Within the RL runs, the checkpoints obtained from the "RLTL;DR, $\lambda=0.01$" and "RLTL;DR, $\lambda=0$" runs should be viewed with caution, as the policy did not learn much and thus had less successful rollouts to learn from. However, "RLTL;DR" and "RLTL;DR, no GRPO loss, only $\mathcal{L}_\text{SFT}$" are comparable since they reach a similar performance in \Cref{tab:sftldr}. Interestingly, just like $\mathcal{L}_\text{SFT}$ alone explained most of the performance of RLTL;DR, it also gives most of the update magnitude. In fact, the magnitude is slightly higher than RLTL;DR, since both are two independent RL runs and $\mathcal{L}_\text{SFT}$-only performed slightly higher, collecting more successful rollouts. Still, it is interesting that $\mathcal{L}_\text{SFT}$-only creates such big updates, although it backpropagates only on 839k tokens compared to the full RLTL;DR's 12M. This shows that the insight internalization indeed is not just a naive next-token prediction but likely activates and updates larger semantically related parts of the network. Note that this cannot be explained by differences between bf16 and fp32 -- inside the RL runs, $\mathcal{L}_\text{SFT}$ is applied by just changing the backprop masks, but the tensors and their precisions follow the mostly bf16 setup of the RL loop. 

Inside the SFT runs, the image is straightforward: SFTL;DR produces a smaller and more concentrated update than SFT on full rollouts, but from \Cref{tab:sftldr} we also know that it trains on fewer tokens and improves performance less. This again points to the fact that, in general, SFT on only the insight tokens is structurally not too dissimilar from SFT on full rollouts. 

\subsection{Performance on revisited tasks}
\label{app:repeat_visits}

\begin{figure}[htbp]
    \centering
    \includegraphics[width=0.55\linewidth]{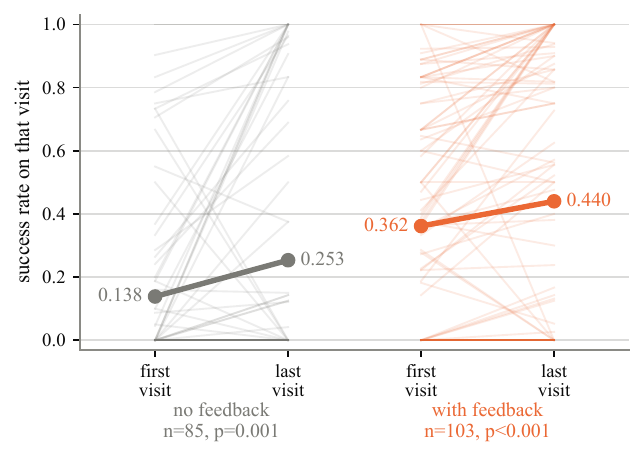}
    \caption{Performance for the same task, revisited multiple times during training again. Left: no
    insight in context. Right: insight in context.}
    \label{fig:learning_patterns}
\end{figure}

Some tasks are sampled more than once during training, allowing us to compare a task's first visit to its last one (\Cref{fig:learning_patterns}). \textit{Without any insight in context}, the optimized policy
solves these tasks nearly twice as often as at the beginning of training ($0.138 \to 0.253$, $n=85$, $p=0.001$). The improvement is about as large as the one the policy makes on tasks it never revisits ($+12.4$pp). The gain thus does not appear to be confined to rollouts with insight in context: what training yields is a general improvement in capability. %

\subsection{GRPO loss is not necessary but speeds up training}
\label{sec:grpo_epoch_benefit}

\begin{figure}[t]
    \centering
    \includegraphics[width=0.72\linewidth]{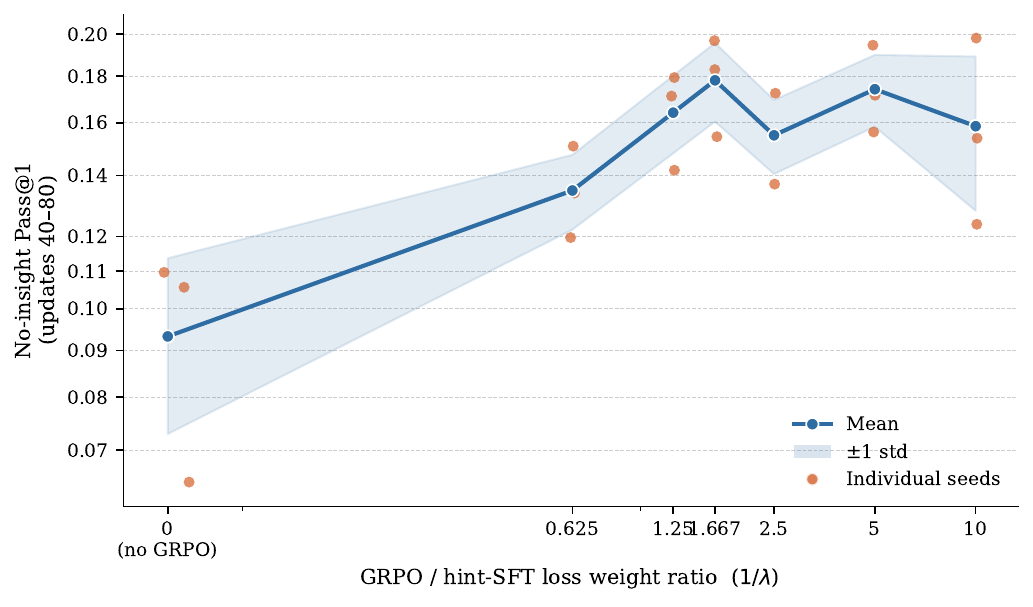}
    \caption{No-insight Pass@1 (averaged over gradient updates 40--80, i.e.\ within the first epoch over the training set) versus the ratio of the GRPO loss weight to the insight-SFT loss weight ($1/\lambda$). Ratio~$=0$ corresponds to the pure SFTL;DR run (no GRPO); ratio increases as $\lambda$ decreases relative to a fixed GRPO coefficient of~1. Gradient clipping (max norm~$=1$) fires on nearly every update regardless of the ratio, keeping the total gradient norm approximately fixed; the ratio therefore controls what \emph{fraction} of this fixed-norm gradient is directed by the GRPO signal versus the insight-SFT signal. Shaded band shows $\pm 1$ std; individual seeds are shown as dots ($n=3$ per point).}
    \label{fig:hint_sft_weight_ratio}
\end{figure}

The fact that RLTL;DR is almost matched by no-GRPO in \Cref{sec:sftldr} is conditioned on training both methods until convergence. The no-GRPO entry in \cref{tab:sftldr} required approximately 25\% more environment interactions than the standard 170k-step budget. Note that this is without additional data, just by increasing the number of epochs from 1 to 1.25. %

To examine whether the additional signal that GRPO gives speeds up convergence, we sweep the GRPO-to-SFT loss weight ratio ($1/\lambda$) across independent runs, using 3 random seeds per ratio value, and report Pass@1 (on rollouts without hints) throughout the mid-early updates $40-80$ in \Cref{fig:hint_sft_weight_ratio}.

Because gradient clipping fires on nearly every update, the total gradient norm remains approximately constant across all ratio values. The ratio $1/\lambda$ thus controls the \emph{direction} of the update, not its magnitude, specifying what fraction steers the policy via task-success signal (GRPO) versus insight internalization (hint-SFT), and we find RLTL;DR to be relatively robust to its choice. The no-GRPO baseline (ratio~$=0$) is consistently outperformed by any run that includes a non-zero GRPO component. This suggests that allocating even a modest fraction of the gradient to the task-success signal accelerates training, though hint-SFT alone accumulates sufficient signal to match full RLTL;DR performance if given more time.

\subsection{Insight evolution over time}
\label{app:diversity}

\begin{figure}[htbp]
    \centering
    \includegraphics[width=0.55\linewidth]{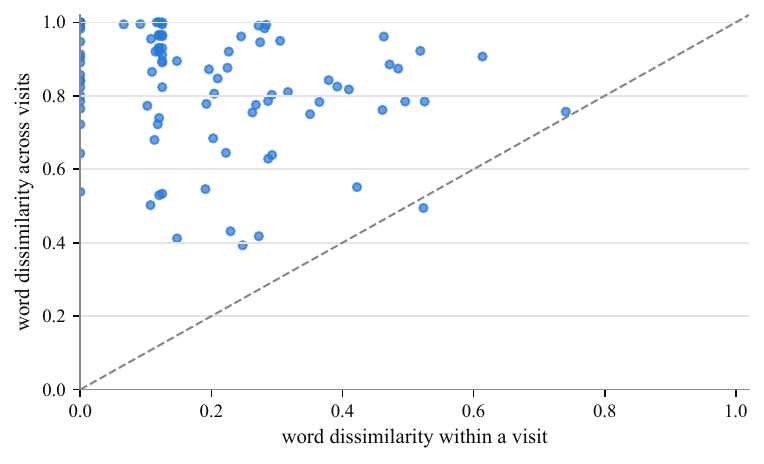}
    \caption{Word-level dissimilarity of the insight a task receives, within one visit (x-axis) against between its first and last visit (y-axis) at different updates. One point per task. Points above the diagonal indicate that insight changes more between visits at different updates than it does between retries at the same update.}
    \label{fig:hint_change}
\end{figure}

\paragraph{Insight evolves with the policy.} A task can be occasionally sampled a few times during training, which lets us ask whether the insight it receives evolved with the policy. We compare the insight written at a task's first visit with the insight written at its last. Similarity is the Jaccard overlap of content words, averaged over sampled pairs of insight strings. Dissimilarity is $1-$ similarity.
Results are shown in \Cref{fig:hint_change}. The x-axis is the mean dissimilarity between two pieces of insight drawn from the \emph{same} visit, which measures how much the verifier varies its wording about a task at one moment; the y-axis is the mean dissimilarity between insight drawn from the two \emph{different} visits.
Every task falls above the diagonal, in the top-left region of the plot, with within-visit dissimilarity averaging $0.19$ against $0.82$ across visits ($n=92$ revisited tasks). The insight a task receives thus varies substantially between visits, consistent with the policy adopting a different strategy as training progresses and the verifier consequently identifying a different type of failure.

\subsection{Insight reliance}
\label{app:reliance}

\begin{figure}[htbp]
    \centering
    \includegraphics[width=0.8\linewidth]{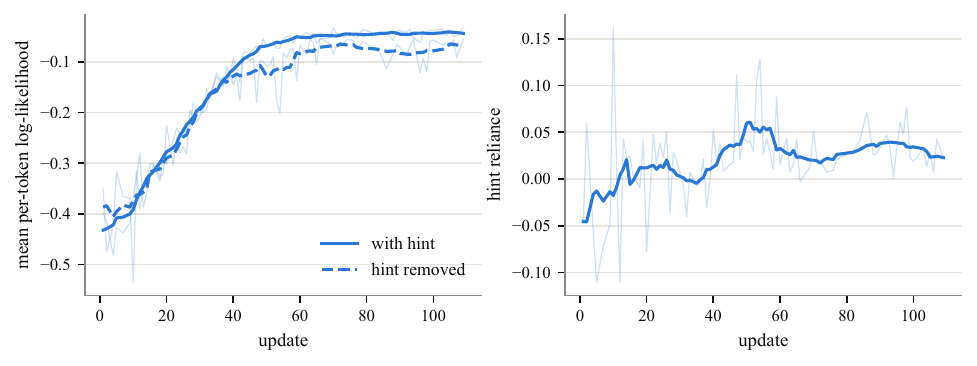}
    \caption{Insight reliance over training. \textbf{Left:} mean per-token log-likelihood of \emph{successful} trajectories, with the insight in context (solid) and with it removed (dashed). \textbf{Right:} their difference, the insight reliance of \citet{xia2026learninghintreinforcementlearning}. Faint lines are per-update values, bold lines a 9-update rolling mean.}
    \label{fig:reliance}
\end{figure}

\citet{xia2026learninghintreinforcementlearning} introduced the notion of \emph{insight reliance}, i.e., $\rho(\tau; q, h) = \log \pi_\theta(\tau \mid q{+}h) - \log \pi_\theta(\tau \mid q)$,
averaged over correct trajectories with insights in context and normalized by trajectory length. This measure should reflect how much a successful trajectory depends on the insight still being present. In their paper, \cite{xia2026learninghintreinforcementlearning} show that low reliance implies successes with insights are more likely to transfer once the insight is removed, and train the insight generator explicitly to keep it low. We measure the same quantity on our run to see
where our insight falls on that scale.

\paragraph{Insight reliance remains mild.} \Cref{fig:reliance} shows reliance rising, but only mildly: from $+0.009$ averaged over the first half of training to $+0.034$ over the second. Some increase in reliance is expected: if the insight raises the success rate at all, reliance cannot be zero. However, reliance remains low in our setup (also considering the numbers reported by \citet{xia2026learninghintreinforcementlearning} without their transfer-weighted reward). The insight is therefore used without being leaned on: it is present in the trajectories the policy learns from, but it is not so load-bearing that such trajectories become implausible once it is removed, consistent with \Cref{app:repeat_visits}.

\subsection{Less difficult Synthetic API split} \label{app:sapi_4pct}

\begin{figure}
    \centering
    \includegraphics[width=0.5\linewidth]{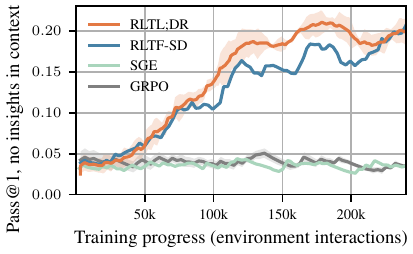}
    \caption{Pass@1 (on rollouts without insight in context) when training on the Synthetic API split with 642 tasks, which includes some solvable tasks so that the baseline Pass@1 is 4\%.}
    \label{fig:sapi_4pct}
\end{figure}

Besides the Pass@128=0 split of SAPI, which contains 458 tasks, we also train on a split which contains 642 tasks. This split came from an earlier Pass@128=0 filtering, where we used not yet optimized sampling hyperparameters. It contains the 458 tasks of the final Pass@128=0 split, plus 184 additional tasks (that with the later improved sampling hyperparameters became solvable in at least 1 of 128 attempts). None of these tasks is trivial, the Pass@1 of Qwen 3.5 9B (with optimized hyperparameters) on the 642 task split is 4\%. It thus gives a good testbed where learning signal is available, if sparse. We present results in this section and note that also the ablations in \Cref{sec:sftldr,sec:abl_feedback} are based on this split. 

\Cref{fig:sapi_4pct} shows that, despite learning signal being present, GRPO and SGE still cannot learn and stay at the original Pass@1 of 4\% of the Qwen 3.5 9B model. RLTL;DR and RLTF-SD both break through the learning barrier.

\subsection{Normal-difficulty dataset splits} \label{sec:exp_fulldata}

\begin{figure*}
  \centering
  \begin{subfigure}[b]{\textwidth}
    \includegraphics[width=\textwidth]{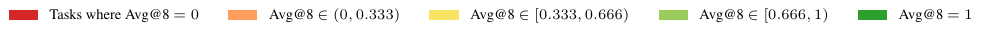}
  \end{subfigure}
  \par\medskip %
  \begin{subfigure}[b]{0.32\textwidth}
    \includegraphics[width=\textwidth]{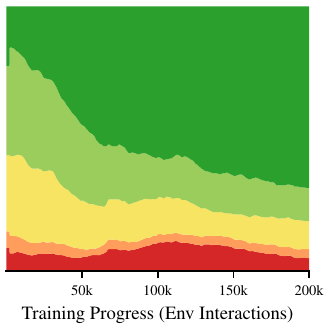}
    \caption{RLTL;DR}
  \end{subfigure}
  \begin{subfigure}[b]{0.32\textwidth}
    \includegraphics[width=\textwidth]{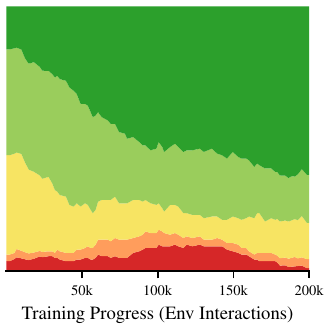}
    \caption{RLTF-SD}
  \end{subfigure}
  \begin{subfigure}[b]{0.32\textwidth}
    \centering
    \includegraphics[width=\textwidth]{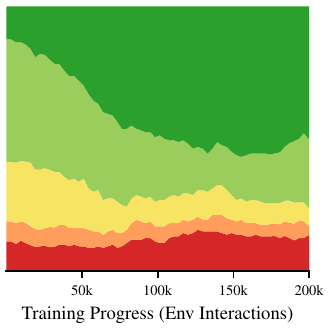}
    \caption{GRPO}
  \end{subfigure}
  \caption{
    Throughout training on the SAPI normal-difficulty split, how many train tasks have a certain success rate. Avg@8=0 means no rollout within 8 was successful (hence  no learning signal), Avg@8=1 means all 8 rollouts were successful (no learning signal either, but well-solved task). Categories in-between are tasks for which the group has a mixture of advantage, hence can learn. \emph{This includes insight-conditioned rollouts, so the plots should be used to compare learning dynamics, not performance.}
  }
  \label{fig:stacked_success}
\end{figure*}

\begin{figure*}[t]
  \centering
  \begin{subfigure}[b]{0.45\textwidth}
    \includegraphics[width=\textwidth]{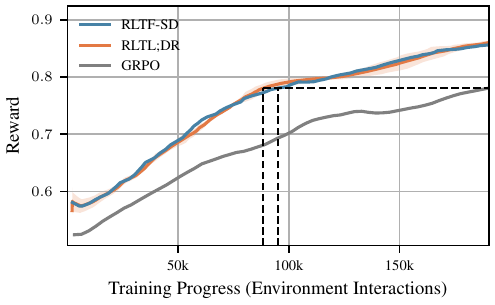}
    \caption{Synthetic API}
    \label{fig:sub1}
  \end{subfigure}
  \hfill
  \begin{subfigure}[b]{0.45\textwidth}
    \includegraphics[width=\textwidth]{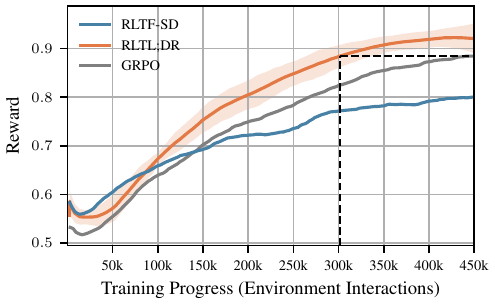}
    \caption{Appworld}
    \label{fig:sub3}
  \end{subfigure}
  \caption{Reward achieved during interactions with the environment. Average and standard deviation across 3 seeds.  We find that RLTL;DR learn more efficiently and achieves more rewards given the same number of interactions due to its richer learning signal and sequential nature of its rollouts.}
  \label{fig:normal_difficulty}
\end{figure*}

\begin{figure*}[t]
  \centering
  \begin{subfigure}[b]{0.32\textwidth}
    \includegraphics[width=\textwidth]{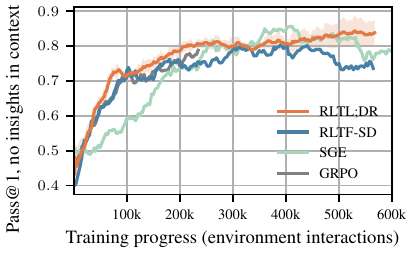}
    \caption{Synthetic API}
    \label{fig:sub11}
  \end{subfigure}
  \hfill
  \begin{subfigure}[b]{0.32\textwidth}
    \includegraphics[width=\textwidth]{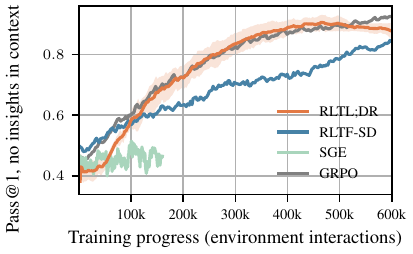}
    \caption{Appworld}
    \label{fig:sub31}
  \end{subfigure}
  \hfill
  \begin{subfigure}[b]{0.32\textwidth}
    \includegraphics[width=\textwidth]{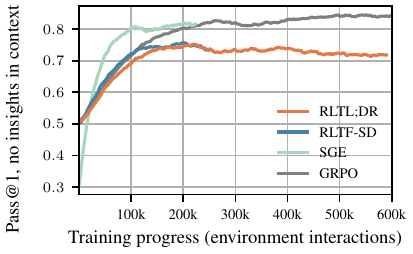}
    \caption{Leetcode}
    \label{fig:sub21}
  \end{subfigure}
  \caption{Pass@1 while training on the full, non-filtered datasets, measured only on rollouts without insight conditioning (comparable between approaches). We find no larger difference between the approaches on (the relatively easy versions of) SAPI and Appworld, but also no drop in performance. On Leetcode, both RLTL;DR and RLTF-SD stagnate after an initial phase of learning. }
  \label{fig:normal_difficulty_env_success}
\end{figure*}

On normal difficulty splits, where $\geq$95-97\% of tasks are solvable in less than 128 attempts, GRPO is able to achieve the same performance as RLTL;DR when sufficiently trained. We do not claim outperformance on such setups. 

To better understand the learning dynamics, we plot the Avg@8 of the tasks during Synthetic API training in \cref{fig:stacked_success}. Note that these include different amounts of insight in the different approaches, so they tell about training dynamics, not performance. 
As depicted in this figure, as training progresses, more tasks are pushed to higher solve rates by RLTL;DR and RLTF-SD while GRPO fails to present the same improvement. In particular, after 100k environment interactions, GRPO does not seem to push the tasks in  the middle solve rates (Avg@8 $\in (0, 0.333)$ and Avg@8 $\in [0.333, 0.666)$) to higher levels while the number of  tasks it always fails on (Avg@8=0) increases. Among the two insight-based methods, RLTL;DR presents a better dynamic  as it consistently maintains a higher proportion of fully solved tasks (Avg@8=1) during the training compared to RLTF-SD. 

\cref{fig:normal_difficulty} shows the reward achieved by different algorithms during training on Synthetic API and Appworld datasets with normal difficulty. According to this figure, RLTF-SD and RLTL;DR perform similarly on Synthetic API dataset and both are consistently achieving higher rewards compared to GRPO during training. Specifically, after only ~91K environment interactions, these two insight-based methods collect the same reward as GRPO collects in 200k interactions, yielding 54\% efficiency that remains even when correcting for the $\sim1.5\times$ higher walltime. %

The pure reward, however, includes rollouts with insights in the context. \cref{fig:normal_difficulty_env_success} shows the Pass@1 on rollouts without insights in context. RLTL;DR and RLTF-SD perform similarly to GRPO on Synthetic API. On Appworld dataset however, RLTF-SD falls behind. On Leetcode, both approaches start like GRPO but then stagnate. We also test OOD performance by evaluating on the unseen Appworld \texttt{test\_challenge} split. GRPO and RLTL;DR produce similarly strong models here. When trained on SAPI, RLTL;DR achieves 52\% Pass@1 and GRPO 53\%. When trained on Appworld-train, RLTL;DR achieves 72.1\% and GRPO 72.2\% Pass@1 on test-challenge.

\subsection{Exploration effectiveness of sequential insight conditioning}
\label{app:inference_curves}

\begin{figure}[H]
    \centering
    \includegraphics[width=0.65\linewidth]{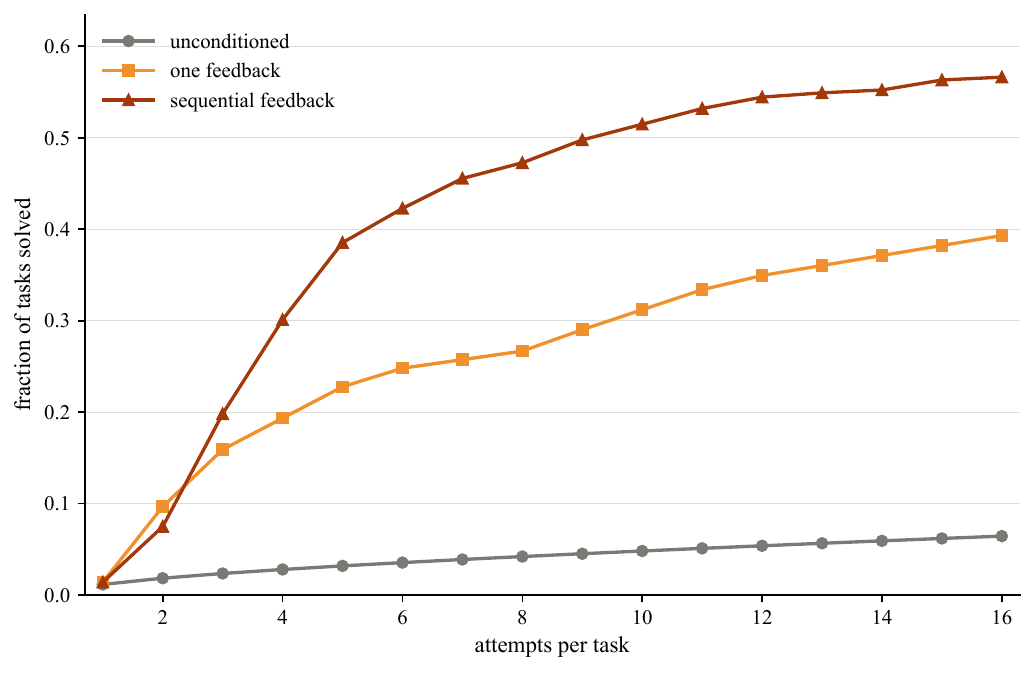}
    \caption{\textbf{Tasks with pass@128$\sim$0.} Pass@k when generating a group of up to $k=16$ rollouts, either with no insight, one insight, or sequential insights, measured with Qwen 3.5 9B Thinking as the rollout and insight policy. Conditioning on previous failed attempts finds a solution for ${\sim}57\%$ of tasks, while a single insight recovers ${\sim}38\%$ and i.i.d. sampling ${\sim}6\%$.}
    \label{fig:passatk128}
\end{figure}

\begin{figure}[H]
    \centering
    \includegraphics[width=0.65\linewidth]{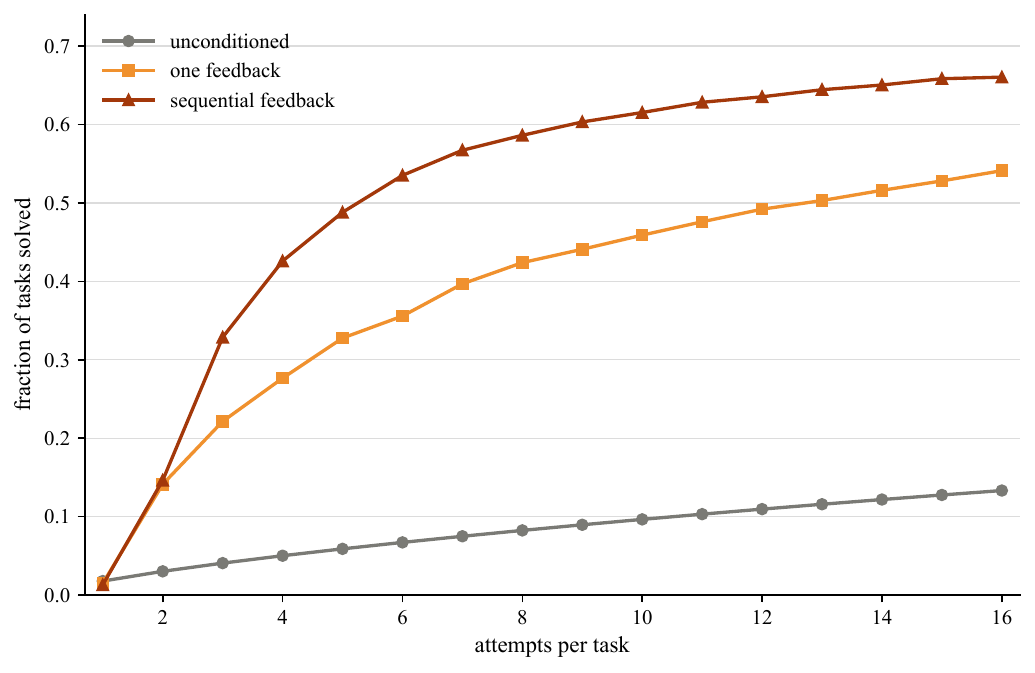}
    \caption{\textbf{Tasks with pass@32$\sim$0.} Pass@k when generating a group of up to $k=16$ rollouts, either with no insight, one insight, or sequential insights, measured with Qwen 3.5 9B Thinking as the rollout and insight policy. Conditioning on previous failed attempts finds a solution for ${\sim}66\%$ of tasks, while a single insight recovers ${\sim}54\%$ and i.i.d. sampling ${\sim}13\%$.}
    \label{fig:passatk32}
\end{figure}

\Cref{fig:passatk128,fig:passatk32} isolate the effectiveness of the sequential insights in finding solutions to very hard problems. They are measured before any training on the base Qwen 3.5 9B Thinking policy. The only thing that changes along the x-axis is how many insights populate the model's context, while the y-axis measures Pass@k. We find that GRPO struggles to find valid solutions, while insight-conditioning elevates performance. The largest gains appear in the first 5 insights. Sequential insight rises faster and higher than just using the latest insight in every attempt. The advantages of conditioning on multiple hints diminish if the policy is trained on such hints, rather than kept frozen.

\subsection{Adding Internalization to RLTF-SD}
\label{app:rltfsd_lsft}

\begin{figure}[H]
    \centering
    \includegraphics[width=0.65\linewidth]{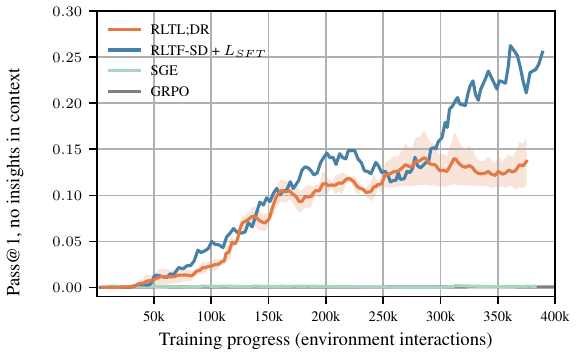}
    \caption{RLTF-SD on SAPI Pass@128=0 when activating the backpropagation masks of $\mathcal{L}_\text{SFT}$.}
    \label{fig:rltfsd_lsft}
\end{figure}

\Cref{fig:rltfsd_lsft} shows that adding $\mathcal{L}_\text{SFT}$ to RLTF-SD, by simply flipping backpropagation masks, might further help performance. We see this as a promising direction for future work, especially since it is simple to implement. One has only to flip the backpropagation mask of the already existing context with insights in it, and, depending on the implementation of the Self-Distillation objective, make sure that the presign is correct so logits of these hints are maximized. 

\section{Understanding and simplifying insight} \label{sec:abl}

\subsection{Training the insight generator versus internalizing insight directly} \label{sec:abl_feedback_gen}

RLTF \citep{song2026rltf}, already introduced as our RLTF-SD baseline (\Cref{sec:exp_frontier}), uses the same sequential RL framework as ours: it conditions each generation round on insight produced from the previous round's rollout. But it also proposes a second variant, RLTF-FM, where it train the process of insight generation itself. That is, they put a loss on the insights $f$ in the context they were generated, i.e. $\pi_\theta(f|\tau)$ conditioned on the actual full rollout (and feedback-generation prompt format), instead of directly learning the mapping from task to insight via the $\mathcal{L}_\text{SFT}$ loss on $\pi_\theta(f|g)$ like in this paper. 

Training the insight generation capability itself is an established idea in the self-play literature, where teaching a model to produce more useful insight, or more relevant follow-up prompts to the original task, has proven effective. \citet{dong2025stp} train a theorem-proving LLM to act as both conjecturer and prover in self-play, rewarding the conjecturer for proposing problems at the edge of the prover's current capability. \citet{liu2025spice} similarly train one model in two roles, a Challenger that mines a corpus to generate reasoning tasks and a Reasoner that solves them, with the Challenger's curriculum adapting to the Reasoner's current skill. We adapt two such approaches to our setting, both with the actor-side $\mathcal{L}_\text{SFT}$ insight loss disabled so we isolate the insight generator's own training signal.

\textbf{RLTF-FM.} Inspired by the RLTF-FM method in RLTF \citep{song2026rltf}, we apply an SFT loss to the insight generator on every piece of insight it produces, regardless of whether the subsequent, insight-conditioned rollout succeeds or fails. This directly supervises the generator to reproduce its own past insight, independent of downstream outcome.

\textbf{Self-play insight SFT.} Aligned with self-play techniques in RL, we instead apply the SFT loss only to \emph{effective} insight -- insight that led the actor to solve the task in the subsequent round, conditioned on it. This steers the insight generator toward generating insight that is useful, rather than merely reproducible.

We find that training the insight generator is not able to substitute for the signal that $\mathcal{L}_\text{SFT}$ on the actor provides in correcting the actor's representations: both variants land close to the GRPO baseline and well below RLTL;DR (\Cref{tab:feedback_generator_training}).

We further test whether insight-generator SFT still helps \emph{on top of} RLTL;DR, i.e., adding it alongside (rather than instead of) the actor-side $\mathcal{L}_\text{SFT}$. We do not see much gain from adding RLTF-FM this way, consistent with its lack of a standalone effect above. Adding Self-play insight SFT, however, gives a more noticeable gain over RLTL;DR alone (\Cref{tab:feedback_generator_training}).

\begin{table}[t]
\centering
\caption{Training success rate on rollouts without insight in context (as in \Cref{tab:feedback-ablation}), averaged over the first 80 gradient updates, on the SAPI-challenging split. Neither way of training the insight generator alone recovers RLTL;DR's actor-side $\mathcal{L}_\text{SFT}$ signal; added on top of RLTL;DR, RLTF-FM gives little further gain, while Self-play insight SFT gives a more noticeable one.}
\label{tab:feedback_generator_training}
\resizebox{\textwidth}{!}{
\begin{tabular}{lc}
\toprule
\textbf{Setting} & \textbf{No-insight success (\%)} \\
\midrule
RLTF-FM (insight generator SFT, all insight) & 6.2 \\
Self-play insight SFT (insight generator SFT, effective insight only)\footnotemark & 7.7 \\
RLTL;DR (ours) & 14.8 \\
\midrule
RLTL;DR + RLTF-FM & 15.0 \\
RLTL;DR + Self-play insight SFT & 15.7 \\
\bottomrule
\end{tabular}
}
\end{table}
\footnotetext{This run's training crashed at update 75, so its value is averaged over the 75 available updates rather than the full 80.}

\subsection{Loss functions and GRPO advantage groups} \label{sec:abl_losses}

In order to make the approach as simple as possible, we have refrained from some technically correct changes to the loss functions in RLTL;DR. In \cref{tab:loss_ablations}, we present ablations. They are trained on the 386 task subsplit of the frontier-difficult SAPI tasks, and evaluated on the remaining 256 tasks, as in \cref{sec:sftldr}.

First, we ablate the decision rule of when to insert insights into the next attempts. By default we do this if the running average success rate in the current batch is $\leq50\%$, to push the batch into a Goldilocks zone. We indeed find that on the very hard tasks, we benefit from more insights, and from backpropagating them with higher $\lambda$: a threshold of 0.5 increases performance over 0.33 and 0.1 would slightly decrease it. If we made our decision rule to let the first 10 rollouts be without insight, then decide for the upcoming (usually 10-11) rollouts to include insight if the first had $\leq3$ successes, it also improves over the other $\lambda=0.1$ runs. This is because net, this would result in more insights, which are generally helpful on very hard tasks.

Second, we test ablations on the GRPO loss, all compared to the $19.6\%$ run. While left unmodified in RLTL;DR for simplicity, one could debias it. We find that splitting the GRPO into two groups, one with the insight-conditioned rollouts and the other without, does not change performance. We leave further exploration of this to future work, since our primary goal is simplicity. Attempting to directly train on the rollouts generated with insights, removing the insights from context at update time (but correcting for this off-policy rollout collection via importance sampling) reduces performance. If this is the goal, we recommend a proper self-distillation loss. Last, adding the GRPO-style division by the std slightly reduces performance, validating our tuning.

\begin{table}[h]
    \centering
    \small
    \caption{Ablations of the RLTL;DR loss. Trained on the 386 SAPI hard tasks, evaluated on the remaining 256 SAPI hard tasks.}
    \begin{tabular}{lc}
    \toprule
       Ablation  & Eval Pass@1 \\
    \midrule
       Insert insights if $\leq 0.1$ running success rate, $\lambda=0.1$ & 14.2\% \\
       Insert insights if $\leq 0.33$ running success rate, $\lambda=0.1$ & 15.3\% \\
       Insert insights if $\leq 0.5$ running success rate, $\lambda=0.1$ & 17.1\% \\
       Insert insights if $\leq$3 of first 10 rollouts successful, $\lambda=0.1$  & 19.2\% \\
       Insert if $\leq 0.5$ running success rate, $\lambda=0.5$  & 19.6\% \\
    \midrule
       Split Advantages & 18.9\% \\
       Off-policy training & 9.6\% \\
       GRPO-style std normalization & 17.0\% \\
    \bottomrule
    \end{tabular}
    \label{tab:loss_ablations}
\end{table}

\end{document}